\documentclass[10pt,journal,compsoc,twoside]{IEEEtran}
\usepackage{hyperref}

\ifCLASSOPTIONcompsoc
  \usepackage[nocompress]{cite}
\else
  \usepackage{cite}
\fi

\usepackage{egweblnk}

\usepackage{hyperref}       
\usepackage{url}            
\usepackage{booktabs}       
\usepackage{amsfonts}       
\usepackage{nicefrac}       
\usepackage{microtype}      
\usepackage{amsmath}
\usepackage{tikz}
\usepackage{float}
\usepackage{siunitx}
\usepackage{enumitem}
\usepackage{graphicx}
\gdef\etal{\textit{et al.}}
\usepackage{xcolor}
\usepackage{gensymb}
\usepackage{amsmath}
\usepackage{tikz}
\usepackage{cleveref}
\usepackage{float}
\usepackage{siunitx}
\usepackage{multirow}
\usepackage{graphicx, tikz,graphics,color, subcaption}
\usepackage{gensymb}
 
\newcommand{\tsmicro}{\omega}
\newcommand{\tsmacro}{\Omega}

\newcommand{\deletethis}[1]{}

\definecolor{brightcerulean}{rgb}{0.11, 0.67, 0.84}
\definecolor{bondiblue}{rgb}{0.0, 0.58, 0.71}
\definecolor{emerald}{rgb}{0.31, 0.78, 0.47}
\definecolor{pigment}{rgb}{0.0, 0.65, 0.31}

\newcommand{\REMOVE}[1]{}

\begin{document}
\title{Neural Photometry-guided \\ Visual Attribute Transfer}

\author{Carlos Rodriguez-Pardo, Elena Garces
	\IEEEcompsocitemizethanks{\IEEEcompsocthanksitem Carlos Rodriguez - Pardo is with SEDDI (28007, Madrid, Spain) and with Universidad Carlos III de Madrid (28005, Madrid, Spain).\protect\\
		E-mail: carlos.rodriguezpardo.jimenez@gmail.com
		\IEEEcompsocthanksitem Elena Garces is with SEDDI (28007, Madrid, Spain) and with Universidad Rey Juan Carlos (28933, Madrid, Spain) \\
		E-mail: elena.garces@seddi.com}%
	\thanks{}}

\markboth{IEEE Transactions on Visualization and Computer Graphics (Pre-Print)}{Rodriguez - Pardo, Garces: Neural Photometry-guided Visual Attribute Transfer}

\IEEEtitleabstractindextext{%
\begin{abstract}
	
We present a deep learning-based method for propagating spatially-varying visual material attributes (e.g. texture maps or image stylizations) to larger samples of the same or similar materials.  
For training, we leverage images of the material taken under multiple illuminations and a dedicated data augmentation policy, making the transfer robust to novel illumination conditions and affine deformations. 
Our model relies on a supervised image-to-image translation framework and is agnostic to the transferred domain; we showcase a semantic segmentation, a normal map, and a stylization. 
Following an image analogies approach, the method only requires the training data to contain the same visual structures as the input guidance.
Our approach works at interactive rates, making it suitable for material edit applications. 
We thoroughly evaluate our learning methodology in a controlled setup providing quantitative measures of performance.
Last, we demonstrate that training the model on a single material is enough to generalize to materials of the same type without the need for massive datasets. 
\end{abstract}

}

\maketitle

\IEEEdisplaynontitleabstractindextext

\IEEEpeerreviewmaketitle

\ifCLASSOPTIONcaptionsoff
  \newpage
\fi

\newcommand{\imacro}{\text{X}}
\newcommand{\imicro}{i}
\newcommand{\lightset}{\text{L}}
\newcommand{\imicroset}[1][]{\mathcal{I}_{\lightset{#1}}}
\newcommand{\map}{\mathcal{M}}

\IEEEraisesectionheading{\section{Introduction}\label{sec:introduction}}

\IEEEPARstart{T}{he} development of effective and editable material models is becoming increasingly important so that users of virtual prototyping, video-games or AR/VR applications can have compelling and realistic experiences. 
For instance, allowing for the creation of virtual environments that surpass the uncanny valley, or empowering artists to create breathtaking visual settings. 
Effective material representations require understanding the properties that uniquely define them, which we refer to as \textit{visual material attributes}. These attributes are spatially-varying parameters that maintain spatial coherency with respect to the material structure,
while remaining invariant to changes in the scene illumination or the geometry of the underlying object.
For example, they may represent optical properties of a microfacet spatially-varying BRDF~\cite{steinhausen2014acquiring} (albedo, normals, roughness, anisotropy, etc.), but also artistic stylizations or higher-level properties, as in semantic segmentation masks.

\begin{figure}[!tb]
	\centering
	\vspace{-0.9cm}
	\includegraphics[width=\linewidth]{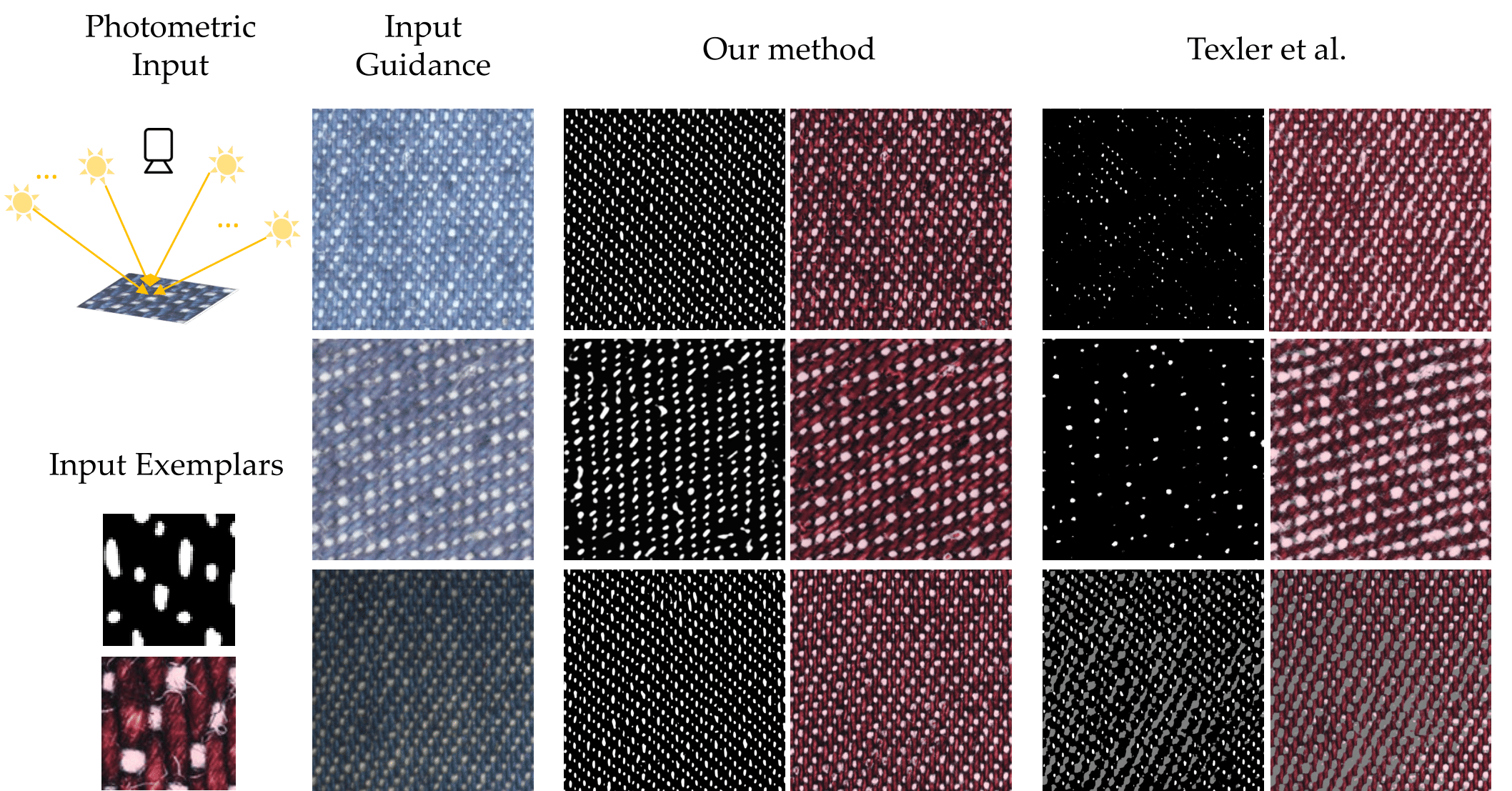}
	\vspace{-0.6cm}
	\caption{The input exemplars on the left-bottom are transferred to the input guidances (second column) using images of the material taken under multiple illuminations (photometric input) as training data. Using this data for training, Texler~\etal~\cite{Texler20SIG} fail to generalize to geometric distortions and illumination conditions not present in the training set.}
	\vspace{-0.8cm}
	\label{fig:teaser}
\end{figure}

Obtaining these attributes for \textit{large} material samples is problematic (\textit{e.g.} requiring large capturing setups or tedious manual input), and can be addressed by estimating these properties in small exemplars and, later, propagating --or \textit{transferring}-- them to the large input image, which serves as guidance. 
This propagation requires adapting to the local and global spatial regularities of the material, which can be challenging if the input guidance image suffers from inhomogeneous illumination, unknown scale, or affine distortions (see Figure~\ref{fig:teaser}, second column).

The problem has been formulated within the context of image analogies~\cite{hertzmann2001image}, and addressed via PatchMatch-based synthesis~\cite{melendez2012transfer}, look-up-tables~\cite{riviere2016mobile}, or neural networks~\cite{mazlov2019neural}, by looking for repetitive patterns, and match color statistics or image gradients. However, most of these methods are prohibitive due to their runtime performance or do not generalize to any kind of visual attribute.
Neural networks, in particular, have proven successful and efficient for the task of video stylization~\cite{Texler20SIG}, where the user inputs a few editing exemplars which are propagated at interactive rates to the rest of the video. However, as shown in this paper, such approach does not generalize to novel illumination conditions.

We propose a novel learning-based method to propagate any visual material attribute --estimated locally for a material-- to larger samples of it. We train a neural network per material using image-to-image translation methods, making use of a policy of data augmentation that makes the transfer invariant to affine transformations (scale, rotations and shears). As opposed to other methods that use synthetic datasets for training and evaluation, our method is robustly tested using a real dataset.
Further, illumination invariance is obtained by feeding the network with multiple images of the material taken under a diverse set of illuminations, which composes our \emph{Photometric dataset}. Our models can be trained in less than a minute and generalize to materials with the same microstructure.

In summary, we present the following contributions: 
\begin{itemize}
	\item The first method to use photometric data to train an image-to-image translation model capable of propagating any kind of visual material attribute to larger samples of the material regardless of the illumination conditions of the input image.
	\item A data augmentation policy, thoroughly evaluated with a real dataset, designed to make the transfer invariant to affine deformations.
	\item Exhaustive comparisons with related work  demonstrating that we can achieve more predictable and higher-quality mappings with a fraction of the computational cost.
	\item Further, we show that our trained models generalize to materials with similar microstructure as the ones used for training.	
\end{itemize}

\section{Related work}
\label{sec:relatedwork}

Several computer graphics and vision problems are closely related to our method. The most similar ones are those related to any form of by-example \textit{visual attribute transfer} (e.g color, texture, style, or geometry). Besides, we also review material estimation and capture methods.

\paragraph*{\textbf{Visual Attribute Transfer}}

Refers to the problem of transferring some visual attributes (e.g. color, style, texture, or geometry) of one or many exemplars to another exemplar while preserving its \textit{content}.

This problem can be formulated within the context of \textit{Image Analogies}~\cite{hertzmann2001image}, in which the goal is to stylize a target un-stylized image B, given a pair of images A (un-stylized) and A' (stylized). The most common approach to tackle this problem has been via patch-based texture synthesis~\cite{benard2013stylizing,barnes2015patchtable,jamrivska2019stylizing}. Nevertheless, recent approaches have leveraged the capabilities of deep latent spaces within convolutional neural networks to disentangle style from content~\cite{reed2015deep}.
A seminal work by Gatys \etal \cite{gatys2015neural} uses a VGG-19 convolutional neural network \cite{simonyan2014very} pre-trained on ImageNet \cite{deng2009imagenet} 
as a feature descriptor for images, in which style and content are related to different layers of the network, and transferred by gradient descent optimization.
Their work on \textit{style transfer} has been extended for single images \cite{huang2017arbitrary, li2017universal, johnson2016perceptual,chen2017stylebank} and video~\cite{chen2017coherent}, as well as for developing image-space distance metrics that resemble human perception \cite{zhang2018unreasonable}. A limitation of many of these methods are their narrow capabilities to provide predictable edits, and considerable focus nowadays is put towards this end~\cite{gatys2017controlling,gu2018arbitrary}. 
A comprehensive review on the topic of neural style transfer is provided by Jing et al.~\cite{jing2019neural}. Our work differs from traditional style transfer approaches in the sense that we deal with a more constrained problem that requires predictable outcomes.

Exploiting the power of deep neural networks in the image analogies problem was tackled by Liao~\etal~\cite{liao2017visual} who, by assuming a semantic prior over an exemplar input image and a target one, propose a method capable of finding a bijective mapping between both inputs, enabling two-way stylizations. Single-image generative models~\cite{shaham2019singan} were extended to the image analogies problem in~\cite{benaim2020structural}, by using convolutional neural networks to generate a new image with the style of an input \textit{style} and the \textit{structure} of another image. These methods, however, rely heavily on content or semantic features, making them vulnerable to lighting or geometric differences between the input images; and are computationally expensive, rendering them impractical for interactive applications. Our method is robust to both geometric distortions and illumination variations, and works at interactive rates. 

Similar in spirit to our method, as it explicitly considers texture variations due to illumination, is the work of Fi{\v{s}}er~\etal~\cite{fivser2016stylit}, which applies patch-match to provide illumination-dependent exemplar-based stylizations to cartoon pictures. In contrast, our work is meant to be illumination-invariant.
Also concerned with stylization problems, Texler~\etal~\cite{Texler20SIG} present a method highly related to ours. They apply patch-based training of an encoder-decoder deep neural network using key frames stylized by a user. Resembling our approach, their algorithm also follows a few-shot learning strategy using as training data a few exemplar patches. However, as opposed to our method, they do not account for the variability of the appearance of materials under different lighting and viewpoint so, as we show in Section~\ref{sec:results}, their method does not generalize to unseen illuminations or geometric variations.

\emph{Image colorization} is concerned with colorizing a gray-scale image given a few colorized exemplars. In this problem, it is critical to infer semantic relationships between the images so that the new scene is perceptually coherent and plausible~\cite{he2018deep, zhang2019deep, he2019progressive}. Similarly, \textit{edit propagation} methods~\cite{an2008appprop,endo2016deepprop} work by propagating strokes provided by a user to the rest of the image, removing the need for a semantic understanding of the input image. Our work is related to the latter techniques, as we perform on the feature spaces of the CNNs and do not require a large labeled dataset to effectively solve our problem, and it can also be used to propagate segmentation masks~\cite{li2008scribbleboost}.

\begin{figure*}[htb]
	\vspace{-0.15cm}
	\includegraphics[width=\linewidth]{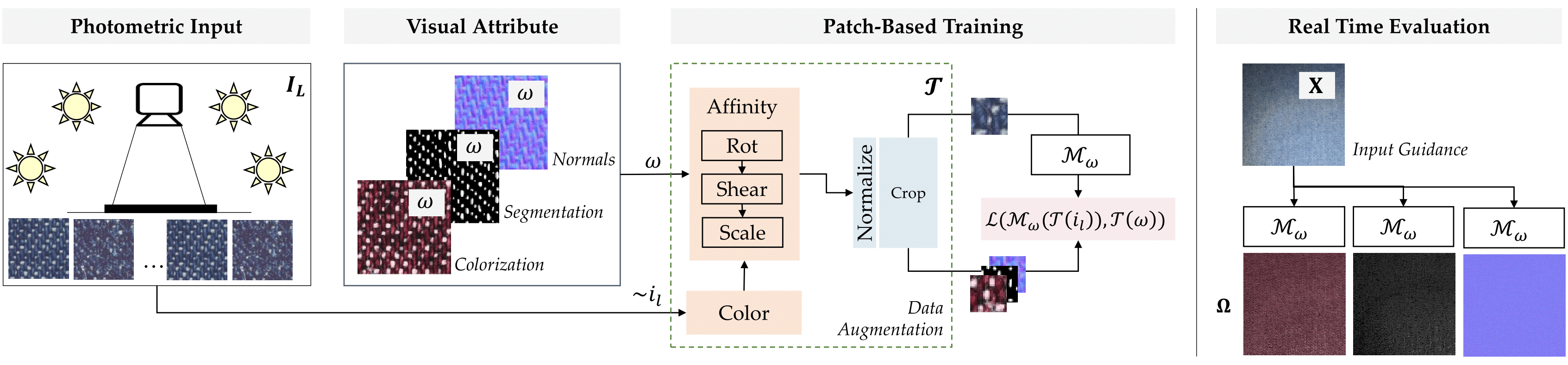}
	\vspace{-0.72cm}
	\caption{Overview of the method. We learn a mapping between the photometric response $I_L$ of the material and a visual property map $\tsmicro$. We make this mapping robust to affine transformations by means of a particular policy of data augmentation used for training. We learn one model $\map$ per material and visual attribute, which allows us to robustly evaluate the performance of the method under several transformations of the guidance images $\imacro$. At evaluation time, $\map$ can have any size. The training of $\map$ per visual attribute $\tsmicro$ takes less than a minute.}
	\vspace{-0.5cm}
	\label{fig:overview}
\end{figure*}

\paragraph*{\textbf{Textured Materials}}

Many real-world materials show spatial regularities, commonly referred to as \textit{textures}. The patterns present in textures can be parameterized, which allows for low-cost material capture or synthesis models. A way to model textured materials is through BTFs (Bidirectional Texture Functions)~\cite{dana1999reflectance,leung2001representing}, a technique that uses multiple camera views and lighting angles to capture a dense sampling of the appearance of a material.
Inspired by such methods, we leverage several images of the material under different illumination conditions, 
however, we require less data than typical BTFs capture setups~\cite{rainer2019neural,rainer2020unified}. 
Similarly, the problem of extrapolating BTFs captures to larger  material samples was addressed by Steinhausen et al.~\cite{steinhausen2015crossdevice,steinhausen2015normals}, who propagate measured BTFs using texture synthesis.
Our method is not meant to propagate full BTFs measurements but could potentially be applied to such datasets, as we illustrate on the supplementary material. 

The goal of texture synthesis is to reconstruct a larger image given a small sample leveraging structural content. This is a long-standing problem in the computer graphics field and different strategies have been proposed, for instance, using PatchMatch \cite{diamanti2015synthesis}, texture transport~\cite{aittala2015two}, point processes~\cite{guehl2020semi,lefebvre2006appearance}, or neural networks~\cite{elad2017style, zhou2018non, fruhstuck2019tilegan, rodriguez2019automatic}. Also related to our work, Li~\etal~\cite{lin2019site} capture the appearance of materials by first estimating their BRDF and, then, synthesizing the high resolution microstructure from a dataset of measured SVBRDFs. Our problem is unlike texture synthesis, as we do not aim to create novel content but to predictably transfer visual material attributes.

\paragraph*{\textbf{SVBRDF Estimation}}

The problem of estimating a SVBRDF model from one or several images using lightweight capture setups is becoming increasingly popular in the literature. Early work~\cite{hertzmann2005example, hertzmann2003shape} leveraged Photometric Stereo~\cite{ikeuchi1981determining} and SVBRDF manifold bootstrapping~\cite{dong2010manifold}
for surface geometry reconstruction, while newer methods exploit the power of deep neural networks.
Recent surveys by Guarnera~\cite{guarnera2016brdf} and Dong~\cite{dong2019deep} contain the most relevant approaches. 
While our method is not meant to estimate the SVBRDF properties of a material, it can be used in combination with those techniques to create larger material assets.

There are a few methods that follow a similar paradigm to ours, transferring pre-estimated SVBRDF maps to a larger material sample. Using PatchMatch texture synthesis, Melendez~\etal~\cite{melendez2012transfer} transfer displacement and albedo maps from small samples of the materials. Their method is limited to daylight illumination and materials present in façades. By means of look-up-tables, and using surface normals and speculars as guidance, Riviere~\etal~\cite{riviere2016mobile} 
transfer surface reflectance captured with controlled LCD lighting to a material sample observed under natural lighting.
Recently, Deschaintre~\etal~\cite{deschaintre2020guided} fine-tune a network trained to estimate SVBRDFs \cite{deschaintre2018single}, to work on larger material samples taking a guidance image as input. 
This approach is limited to transfer a pre-defined set of property maps while our method can transfer any kind. 
The strategy of using multiple images of the material under different illuminations as input data to is not new. Li~\etal~\cite{li2017modeling} and Ye~\etal~\cite{ye2018single} utilize a self-augmentation strategy to make the estimation of the SVBRDF more robust to unknown environment illumination. We are inspired by these approaches to increase robustness in the model predictions.

\section{Problem Formulation}
\label{sec:framework}

Our goal is to transfer a $D$-dimensional spatially-varying \textit{visual attribute} $\tsmicro$ of a material (for example, estimated locally at high resolution) to a larger sample of it.

We formulate the problem with an image-to-image translation approach. 
For training, our method takes as input: a \textit{photometric} dataset $\imicroset$, and a \textit{visual attribute} $\tsmicro$. The \textit{photometric} dataset consisting of a number of RGB planar images of a material, $\imicroset = \{ \imicro_l | \imicro_l \in \Re^{n \times m\times 3} \}$, $| \imicroset | \geq 1$, illuminated with different light sources $l \in \lightset$, of $n\times m$ pixels. This kind of images can be either captured with specific devices~\cite{nam2016simultaneous,merzbach2017high,photoptics19}, or synthetically rendered given an inverse material estimation pipeline~\cite{guo2020materialgan, deschaintre2020guided}. 
The visual attribute being a spatially-varying map $\tsmicro \in \Re^{n \times m\times D}$ of any kind, and dimensions $D$, that maintains pixel-wise correspondence with the photometric input images $\imicroset$. 
Figure~\ref{fig:overview} and Figure~\ref{fig:dataset} show examples of these images for three different visual attributes: a stylization, a segmentation, and a normal map. 

Given this data for a single material, and a strategy of patch-based training,
we learn a function $\map$ that can be applied to a new \textit{guidance} image of the input material (or a similar one) $\imacro \in \Re^{N \times M\times 3}$ of any size $N \times M$, to get its corresponding visual attribute $\tsmacro \in \Re^{N \times M \times D}$:
\begin{align}
	\map: \imacro &\rightarrow   \tsmacro, \\
	s.t. \; \imicro_l &\overset{\map}{\rightarrow} \tsmicro, \; \forall \imicro_l \in \imicroset .
\end{align}

At evaluation time, the guidance image $\imacro$ might contain different colors, scales, illumination, or affine distortions than the images used for training. Figure~\ref{fig:overview} shows an overview of the training and evaluation processes. In Section~\ref{sec:experiments}, we evaluate the conditions of the input guidance image upon which the method provides robust estimations.

\section{Learning Framework}

In this section, we describe the patch-based and data augmentation strategies used for training, the neural network design and loss functions, and the implementation details.

\subsection{Patch-based Training} \label{sec:invariances}
Each training step takes as input a pair of corresponding image patches taken from the photometric input $\imicroset$, and the visual attribute $\omega$. 
Using this data alone already provides a good starting point for generalizing to unseen illumination setups. However, it is not sufficient in scenarios in which the guidance image contains variations due to image noise, a different scale, or any other affine distortion.
In order to make the transfer invariant to these transformations, the network needs to be trained with the appropriate data. 
Data augmentation strategies are essential for reducing the amount of necessary data for training~\cite{shorten2019survey,karras2020training}, however, random strategies not taking into account the particular domain might degrade the quality of the prediction. 
We therefore follow a pre-defined data augmentation policy $\mathcal{T}$ (illustrated in Figure~\ref{fig:overview}), where random operations are performed sequentially.

\begin{figure}[tb]
	\vspace{-0.1cm}
	\includegraphics[width=\linewidth]{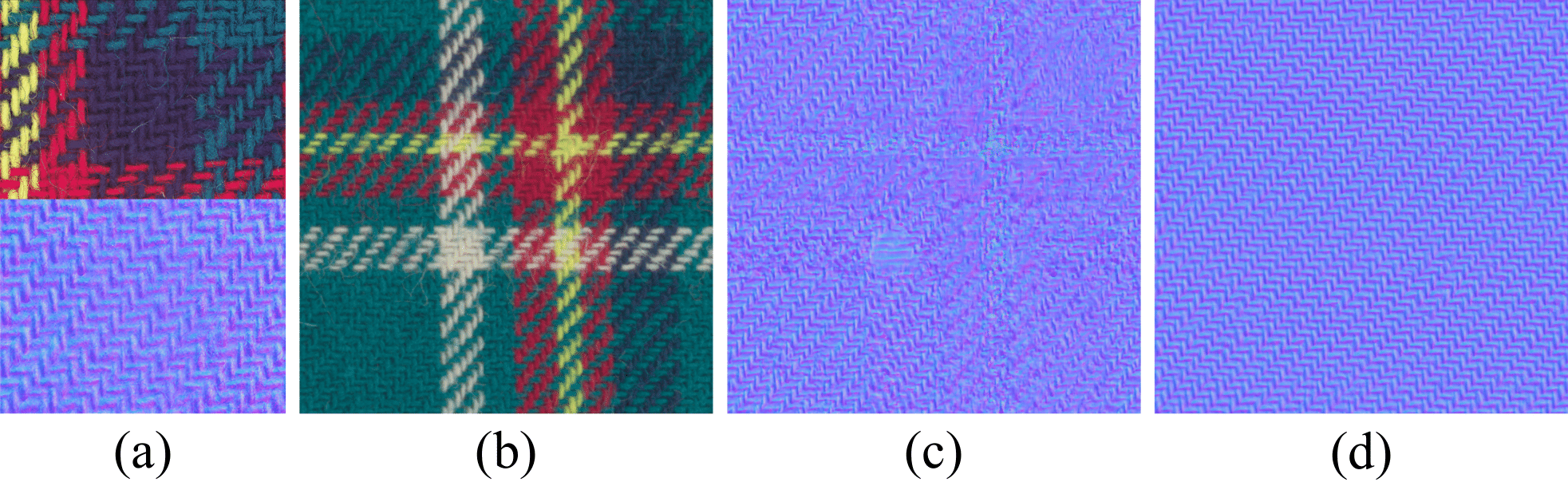}
	\vspace{-0.7cm}
	\caption{The color augmentation policy takes advantage of the material structural regularities to make the transfer robust to different albedos. (a) A diffuse image of the portion of the material used for training, and its corresponding normal map. (b) Input guidance image. (c) Transferred normals without the color augmentation policy. (d) Transferred normals using color augmentation. Note that the training data does not include images containing the white yarn.}
	\vspace{-0.7cm}
	\label{fig:coloraugmentation}
\end{figure}

\paragraph*{\textbf{Color Augmentation}} 
Even if the microstructure of the material is homogeneous and can be measured using only a small patch of it, it may not be possible for the network to estimate its visual attributes for parts of the material which contain previously unseen colors. We correct this by randomly permuting the color channels of the photometric input (see Figure~\ref{fig:coloraugmentation}). As we show in Section~\ref{sec:results}, this data augmentation policy also helps make the model generalize the transfer to similar materials, by learning features that are more related to the structure of the material than to its color. Similar operations have been recently proposed for finding robust visual representations on self-supervised settings~\cite{chen2020simple}. Only the photometric input is subject to this transformation. 

\paragraph*{\textbf{Affine Transforms}}
In order to allow for material editing applications in real images, the model should generalize to images taken under camera perspectives and geometry distortions different than those present in the fronto-planar images used for training.
Many of such texture irregularities can be defined as an affine transform: translation, rotation, shear, or scaling.
As CNNs are shift-invariant by design~\cite{kauderer2017quantifying}, we propose to augment our datasets with random transformations for scalings, rotations, and shears. Those transforms can be efficiently performed to images through matrix multiplications. 
However, some visual property maps need to be treated specially, as the spatial transform might have a different behavior in $3D$ vector space, e.g. normals or tangents. In those maps, each pixel is a representation of a $3D$ vector. As such, we also perform the rotations and shear operations to these maps in $3D$ space, by multiplying each normal vector by the same affine transformation matrix applied to the $2D$ image. We perform the rotation around the $Z$ axis, thus, assuming the camera sensor is parallel to the object plane.

\paragraph*{\textbf{Cropping}} Inspired by recent work on patch-based learning~\cite{Texler20SIG, park2020contrastive}, during training, the network receives small patches of each input pair. Those patches are randomly cropped from the randomly augmented images, so the network receives a considerable amount of variations of the same material, thus making generalization possible.

\begin{figure*}[tb!]
	\vspace{-0.1cm}
	\includegraphics[width=.98\linewidth]{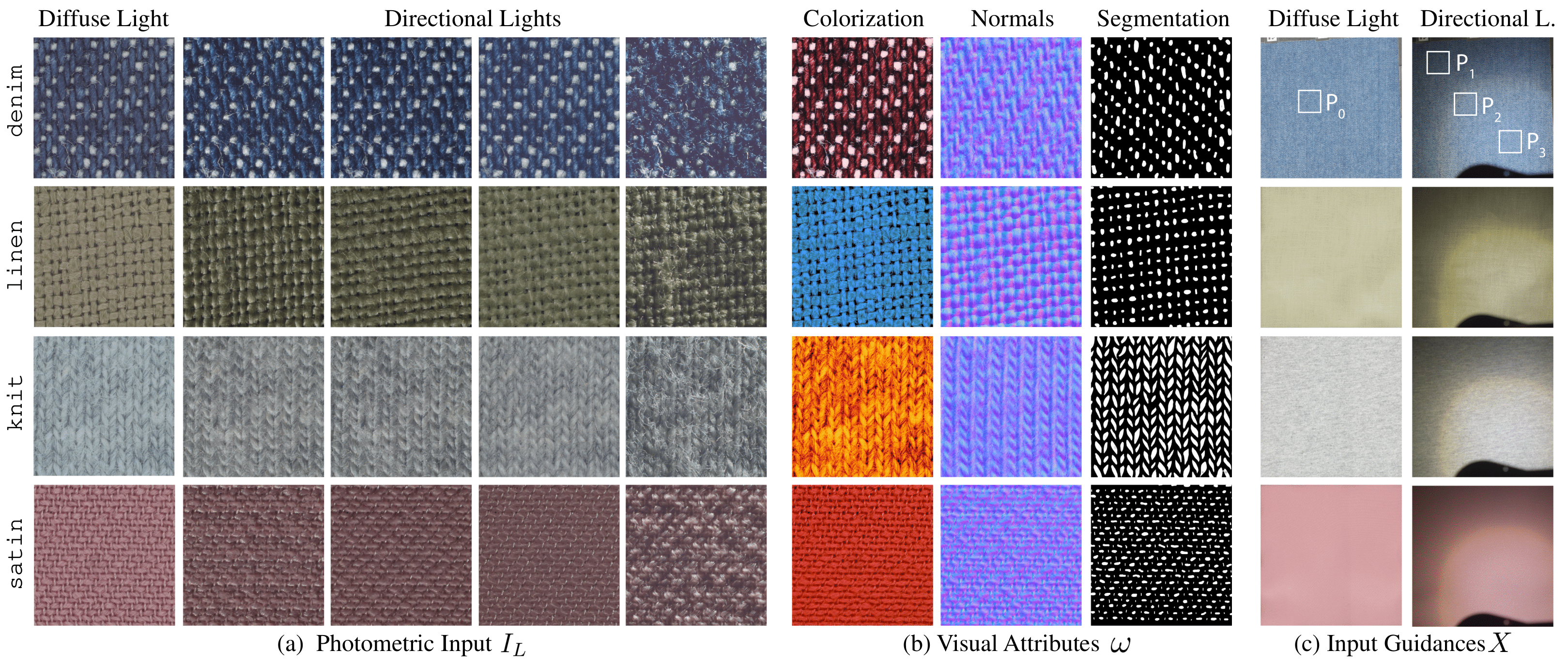}
	\vspace{-0.4cm}
	\caption{An overview of our evaluation dataset. (a) Five example images of our high resolution captured data illuminated with diffuse light and four directional light sources. (b) Examples of some of our \emph{visual property maps}: colorization, normals and yarn segmentation. (c) Evaluation images taken under diffuse and directional illumination sources. \textit{Denim}, ~(c) contains labels for several patches used for the evaluation in Section~\ref{sec:experiments}.		
		Those materials show different properties that may prove challenging for our visual attribute transfer task. Our \emph{denim} and \emph{knit} fabrics show diverse color variations (different colored dyed yarns and stochastic albedo, respectively), \emph{linen} shows strong geometric variations and \emph{satin} shows anisotropic optical behavior. High-resolution copies of these images are included in the supplementary material. }
	\vspace{-0.45cm}
	\label{fig:dataset}
\end{figure*}

\subsection{Network Design}

We follow a uni-modal image-to-image translation learning strategy, assuming there is only one correct mapping from input to output image. This approach is reasonable for the kind of transfers we test in this paper. However, it might fail for ambiguous cases where there are multiple suitable outputs for the same input, for which multi-modal approaches \cite{zhu2017toward} or GANs are more advisable although harder to train.
Specifically, our model $\map$ is a shallow U-Net network~\cite{ronneberger2015u} with 4 blocks of layers, containing a small number of trainable parameters, inspired on few-shot learning strategies~\cite{wang2020generalizing,wang2019few,liu2019few}. This type of fully convolutional architecture is of common usage in image-space regression problems, due to its capability of efficiently learning patterns at different levels of abstraction thanks to its multi-scale design. Skip connections are added to enhance local details~\cite{ronneberger2015u,isola2017image}. The small number of parameters allows for faster training and inference, as well as reduced memory usage. 
We include further details, analysis and discussion in the supplementary material.

Note that, as opposed to previous work~\cite{deschaintre2020guided} on material transfer, that is initialized from a pre-trained network, but inspired by single-sample image synthesis methods~\cite{zhou2018non, shaham2019singan}, we train one network per material and visual property. This strategy, although increasing training time, is key for the following reasons: First, it allows us to better understand the generalization capabilities obtained through our data augmentation policies. Second, it guarantees predictability of the trained model as every feature learned by the network is specific for each material and visual property pairs. Finally, it removes potential problems of a biased dataset as there is no cross-material or cross-domain learning. If the input dataset does not contain enough variations of the material to represent the whole material, the model will fail on areas with unseen patterns. In those cases, having a pre-trained network may help as a material prior, as in~\cite{deschaintre2020guided}. However, as we show on the supplementary material, our method obtains comparable results to pre-trained methods, with a smaller computational footprint and with the additional flexibility of not needing an expensive dataset and large models. In practice, training a single network per material and attribute is not critical as this process takes less than a minute.

\paragraph*{\textbf{Loss function}}
 Choosing the appropriate loss function for a learning framework highly depends on the problem. For example, some methods~\cite{isola2017image} combine a per-pixel $\ell_1$ metric with adversarial~\cite{goodfellow2014generative} losses, as the latter allows for better semantic mappings in multi-modal learning scenarios, whilst the former allows for improved predictability. 
Texler~\etal~\cite{Texler20SIG} further includes a perceptual loss~\cite{texler2020arbitrary},
while using a render loss is common in methods that estimate material parameters from photos~\cite{deschaintre2018single,deschaintre2020guided}. 
In our method, we show that using a $\ell_1$ loss for training is enough for learning accurate and predictable mappings in our regression task, and binary cross-entropy loss for semantic segmentation, following standard practice in image segmentation~\cite{ronneberger2015u}. We found that the $\ell_2$ loss function yields overly-smooth outputs, and perceptual loss functions like \textit{LPIPS}~\cite{zhang2018unreasonable} are more prone to artifacts than pixel-wise $\ell_{norm}$ metrics. We include an ablation study of the impact of the loss function in the supplementary material.

\subsection{Implementation details}  \label{sec:implementation}

We use PyTorch~\cite{paszke2017automatic} as the learning framework, Adam~\cite{kingma2014adam} for optimization, a learning rate of $0.002$, and a batch size of $16$. 
The training images are randomly augmented using uniform distributions by the following operations, in order: First, the photometric input is subject to the color augmentation policy. Then, both inputs and targets are randomly rotated by an angle in the $[-90, 90]$ range, randomly sheared by an angle in the $[-45, 45]$ range, and randomly rescaled in the $[0.5,2]$ range of scale factors. Then, patches of $128\times128$  pixels are randomly cropped during training to generate a large data-set of images. 

All inputs are always standardized using their own mean and variance.
Each $\map$ is trained for $1000$ iterations, which takes around $1$ minute on a single Nvidia 1080Ti GPU. Due to the fully convolutional nature of our models and their reduced number of trainable parameters,
the guidance images $\imacro$ used for evaluation can be of arbitrary dimensions. We used images of up to $5000\times5000$ pixels for which evaluation takes around $150$ms. 
We refer the reader to the supplementary material for a comprehensive description and a diagram of this model, as well as further implementation details.

\begin{figure*}[ht]
	\vspace{-0.1cm}
	\includegraphics[width=\linewidth]{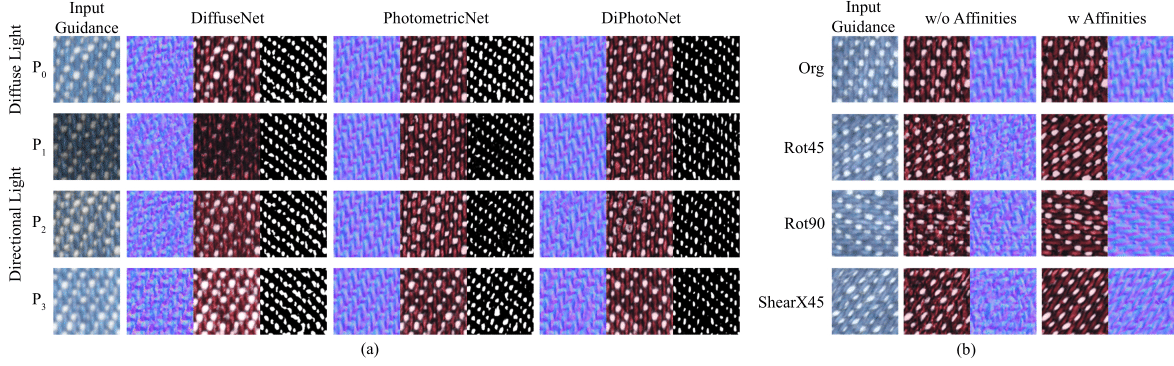}
	\vspace{-0.8cm}
	\caption{Qualitative results of our models under different datasets and data augmentation configurations, for different inputs of the \emph{denim} material. On the left (a), we show the results of our networks trained using different dataset configurations, under a guidance image taken with diffuse lighting ($P_0$), and three crops of a guidance image illuminated with a directional light ($P_1, P_2, P_3$) not present in the training set. Please refer to Figure~\ref{fig:dataset} for the position of these crops on the larger guidance images. 
		On the right (b), we show the results of our \textit{photometricNet}, under different geometric distortions (rotations and shears) performed to its guidances images, taken under diffuse illumination.
}
	\vspace{-0.4cm}
	\label{fig:experiment1}
\end{figure*}

\section{Dataset and Metrics}

\subsection{Dual-Resolution Captured Data}\label{sec:dataset}

Our method is agnostic to the capture setup~\cite{nam2016simultaneous,merzbach2017high,photoptics19}, and it may work for any kind of input data (e.g. BTFs\cite{rainer2019neural,leung2001representing}) as long as the photometric images are pixel-wise aligned with each other and to the visual attribute.
This dataset could also be created synthetically by rendering the outputs of any SVBRDF estimation method~\cite{deschaintre2020guided}. In Section~\ref{sec:results}, we show results of our method using these acquisition pipelines. 

However, for the purpose of this evaluation we chose to work with real data. The main reason is that the data obtained with material capture devices poses extra challenges that are difficult to reproduce with render engines. For example, material irregularities, complex optical behavior, or distortions and color shifts introduced by the optical system that might cause the models to produce inaccurate estimations.

We create a dataset containing images of the same material taken with two imaging camera systems: a high-resolution camera that allows us to take pictures of $0.7\times0.9$~cm, with a resolution of $367\times490$ pixels and a macroscopic camera which provides images of $11\times11$~cm, with a resolution of $4800\times4800$ pixels.
In terms of illumination, our setup has 27 different collimated light sources uniformly distributed across the hemisphere as well as diffuse illumination. 
We build a dataset of four different textile materials, whose complex optical behavior due to anisotropy, transmittance, directionality and microstructure~\cite{castillo2019recent} turns them particularly challenging for synthesis and editing operations~\cite{kampouris2016fine, rainer2019neural}. For training, we capture one image for each light source, making a total of 28 different images (Figure~\ref{fig:dataset}~(a)). For the evaluation set, we take one \textit{guidance} image with diffuse illumination and another \textit{guidance} image with a directional light source (Figure~\ref{fig:dataset}~(c)). The visual attributes (Figure~\ref{fig:dataset}~(b)) are generated automatically using photometric stereo~\cite{ikeuchi1981determining} in the case of normal maps and manually by artists in the cases of colorizations and segmentation masks.

\subsection{Attribute-specific Metrics for Evaluation}\label{sec:metrics}

For evaluating our models, we choose
domain-specific distance metrics different to those they were trained with to better understand their generalization capabilities~\cite{theis2015note}. For normal maps, we compute the cosine distance between ground truth and estimated maps, as it accounts for the geometric space in which normals lie. In the case of image segmentation, we evaluate the results using the Jaccard similarity coefficient (IoU)~\cite{zhou2019context}, which is well-suited for sparse segmentation tasks. 
Finally, we use the state-of-the-art metric \textit{Learned Perceptual Image Patch Similarity} (LPIPS) \cite{zhang2018unreasonable} to evaluate the quality of the colorizations, as it has been shown to outperform $\ell_2$ norm for visual perception tasks.

\section{Evaluation and Comparisons}\label{sec:experiments}
We evaluate our method in different settings. First, we assess the type and amount of photometric input data necessary for the model to generalize to different illumination conditions and image distortions. Then, we test our data augmentation strategy for affine transformations.

\subsection{Invariance to Input Illuminations and Distortions} 

In this set of experiments, we aim to evaluate if the model produces the same output after changing the training data and input guidance images. To this end, we measure both the impact of different illuminations and sizes of the photometric dataset, as well as variations of the input guidance images. 

\subsubsection{Photometric Input} 

In this experiment, we study the type of photometric input that makes the method invariant to different illumination conditions of the guidance image. We compare three models trained with different datasets: \textit{diffuseNet}, which uses a single image illuminated with diffuse lighting; \textit{photometricNet}, which takes as input 27 directional lights; and \textit{diphotoNet}, which uses all the 28 sources.
For evaluation, we use the macroscopic camera which captures a larger sample of the material at a lower resolution. We take two test guidance images with diffuse and directional lighting (Figure~\ref{fig:dataset}~(c)). In this experiment, all images are aligned,
therefore, we follow a limited policy of data augmentation from Section~\ref{sec:invariances}, applying only color augmentation, rescaling, and crops, and leaving out rotations and shears for in-the-wild scenarios.

Figure~\ref{fig:experiment1} shows qualitative results of our method for a selection of patches for the \emph{denim} material. 
Figure~\ref{fig:experiment1}~(a) shows that the best accuracy is in general obtained with \textit{diphotoNet}, i.e. training with both directional and diffuse illuminations. This is reasonable, as the model is trained using the same illuminations used during test time. Conversely, \textit{diffuseNet}, a network only trained using diffuse lighting is less capable of generalizing under any kind of illumination source.
This suggests that following a photometric approach for training material synthesis models allows for better generalization capabilities. 
Finally, the results for \textit{photometricNet}, which is trained only with directional lights shows similar accuracy as \textit{diphotoNet} while proving generalization to unseen illuminations. 
For all the results shown in the paper, we have chosen \textit{photometricNet} as our model. 
This has an additional advantage from the usability perspective: it is relatively easy and cost-effective to build a capture setup (such as the flash light from a smartphone) or generate renders with directional lights, while it is considerably harder to recreate both types of illumination consistently.

\subsubsection{Dataset Size Influence}

Our goal in this experiment is to understand how many images~\cite{nielsen2015optimal} taken under directional lights are needed in order to obtain the desired invariance to illumination.
\deletethis{Measuring quantitative errors in this setting is particularly challenging, as obtaining large-scale material property maps can be prohibitively expensive. However, we find it critical to evaluate each of the decisions taken, including the optimal number of directional lights for training and the data augmentation strategy. In view of the qualitative results of the previous experiment and inspired by recent recommendations for semi-supervised problems~\cite{oliver2018realistic}, we use as ground truth data the transferred results for \textit{photometricNet} evaluated on the guidance image taken under diffuse lighting. By evaluating the models on an image with illumination that was not included in the training data, our confidence on the generalization capabilities of our model on out-of-distribution data examples strengthens. 
}

\begin{figure}[tb]
	\includegraphics[width=1.02\linewidth]{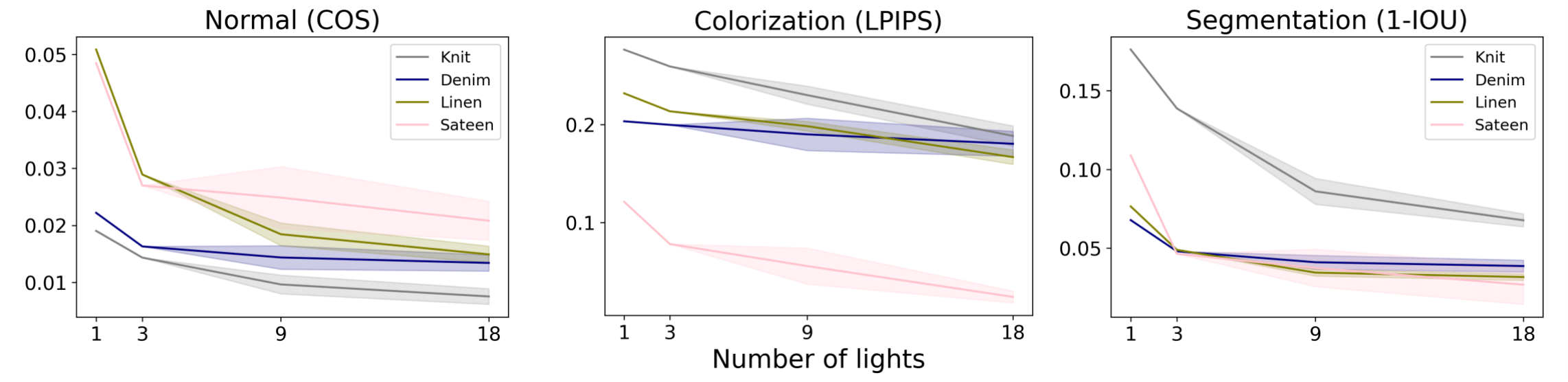}
	\vspace{-0.7cm}
	\caption{Error of the reduced \textit{photometricNets} on our ground truth guidance image (taken with diffuse illumination, not present in the training dataset) for each material in the dataset and the three visual attributes  and corresponding error metrics, which are close to the normal of the surface in which the material lies. The width of the lines indicates the standard variation across 5 repeated experiments, where different light directions were randomly chosen to form the training datasets. Please refer to the supplementary material for the position of each light source. }
	\label{fig:experiment2}
	\vspace{-0.25cm}
\end{figure}

We train different models using reduced versions of our full 27-image dataset, and compare their results to those of the \textit{photometricNet} trained on the full directional dataset. More precisely, we train networks using 1, 3, 9 and 18 directional lights for each material and application in our dataset following the same reduced data augmentation policy described in the previous experiment.
Instead of randomly selecting light sources around the hemisphere, we perform a more sensible light source sampling. Specifically, for each reduced dataset, there is at least one light that is as close as possible to the normal of the surface in which the material lies on, thus giving more importance to frontal angles. We extend these selected lights to 3 for the reduced datasets with more than three lights. 
The rest of lighting sources are uniformly sampled around the hemisphere, up to a zenith angle of 70 degrees. 
The supplementary material contains a diagram of the distribution of lights in the hemisphere.
Results are shown in Figure~\ref{fig:experiment2}, where we see that adding more lights to the dataset monotonically increases the generalization of every model. However, adding lights consistently shows diminishing returns, which suggests that a capture setup with around nine lights might be enough for relatively accurate estimations. Similar findings are reported in multi-image SVBRDF estimation methods~\cite{guo2020materialgan,deschaintre2019flexible}.

\subsubsection{Image Degradations}
\begin{figure}[tb]
	\vspace{-0.1cm}
	\includegraphics[width=\linewidth]{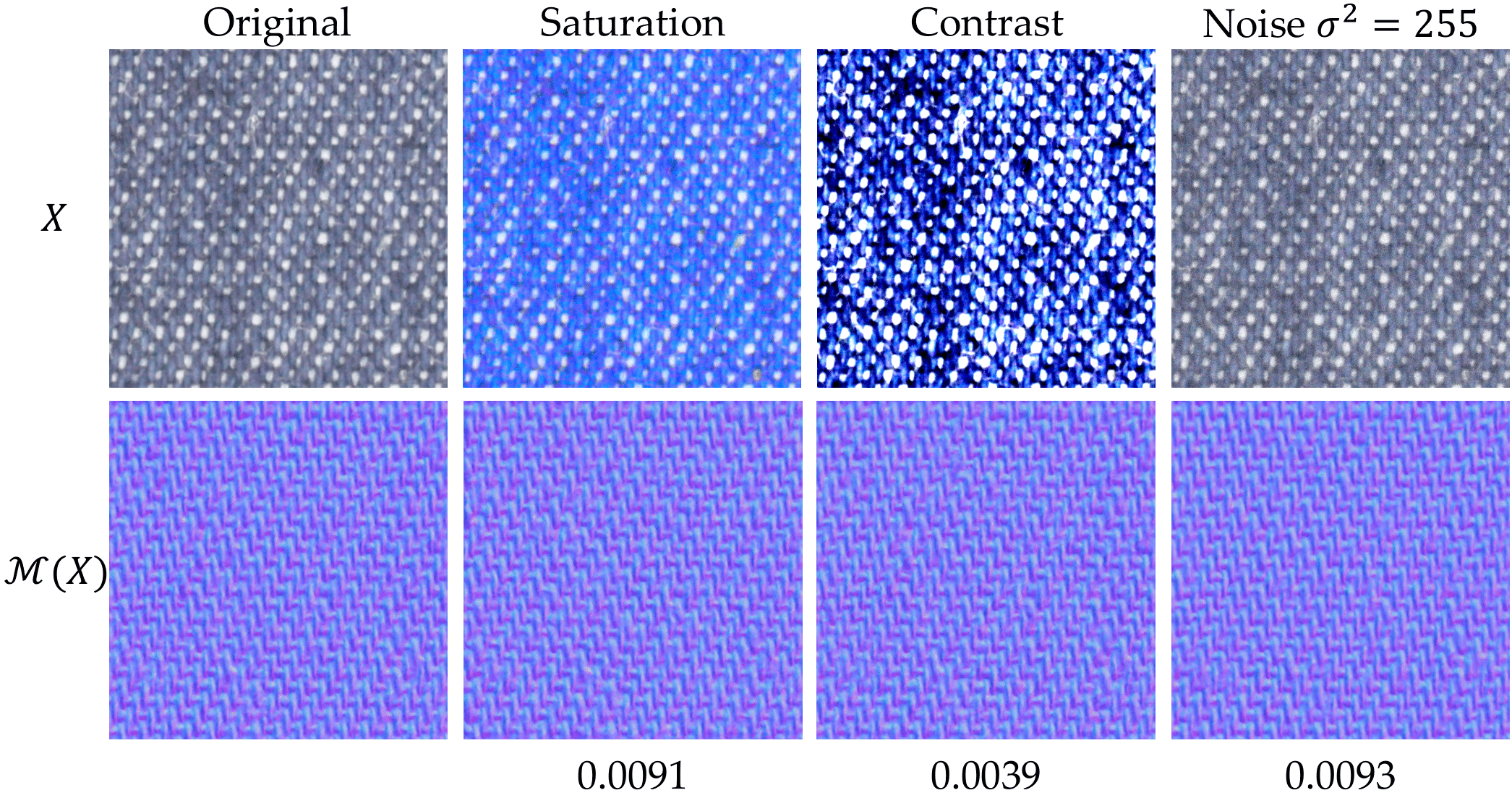}
	\vspace{-0.4cm}
	\caption{Output of our model under different distorted inputs: a change of saturation, contrast, and Gaussian noise ($\sigma^2=255$). The number at the bottom is the cosine distance with respect to the original estimation. As we can see, the output is consistent with a very small error in all cases. 
}		
	\vspace{-0.3cm}
	\label{fig:distortions}
\end{figure}

As discussed in Section~\ref{sec:dataset}, real captured images may be subject to distortions, shifts and noise introduced by the optical capture system. To fully understand the robustness of our models with respect to these types of imperfections, we synthetically modify the saturation, contrast and noise present in the $\imacro$ guidance images. As shown on Figure~\ref{fig:distortions}, $\mathcal{M}$ is robust to these types of degradations, even when a fair amount of details are lost in $\imacro$. We refer the reader to the supplementary material for further examples of these analyses and implementation details. These images were not part of the training dataset.

\subsection{Equivariance to Affine Transforms}

Our goal in this experiment is to analyze the equivariance of our model to affine distortions of the guidance images. A function $f$ is said to be \textit{equivariant} with respect to a transformation $\mathcal{T}$ if $f(\mathcal{T}(x)) = \mathcal{T}(f(x))$. Inspired by~\cite{benton2020learning}, we measure the equivariance of our model $\mathcal{M}$ with respect to different affine transforms $\mathcal{T}$ performed to an input guidance image $\imacro$, by computing their difference using the corresponding metric defined in Section~\ref{sec:metrics} $d(\mathcal{M}(\mathcal{T}(\imacro)), \mathcal{T}(\mathcal{M}(\imacro)))$.   
To achieve this, we extend the augmentation policy in the previous experiments by adding random shears and rotations to the training process.
We train the \textit{photometricNet} with and without random shears and rotations for every visual attribute in our dataset and on the \textit{denim} material. In the case of the normals, as described in Section~\ref{sec:invariances}, we also perform the shears and rotations in the geometric space in which normals lie.

 \begin{figure}[!tb]
	\centering
	\vspace{-0.1cm}
	\includegraphics[width=\linewidth]{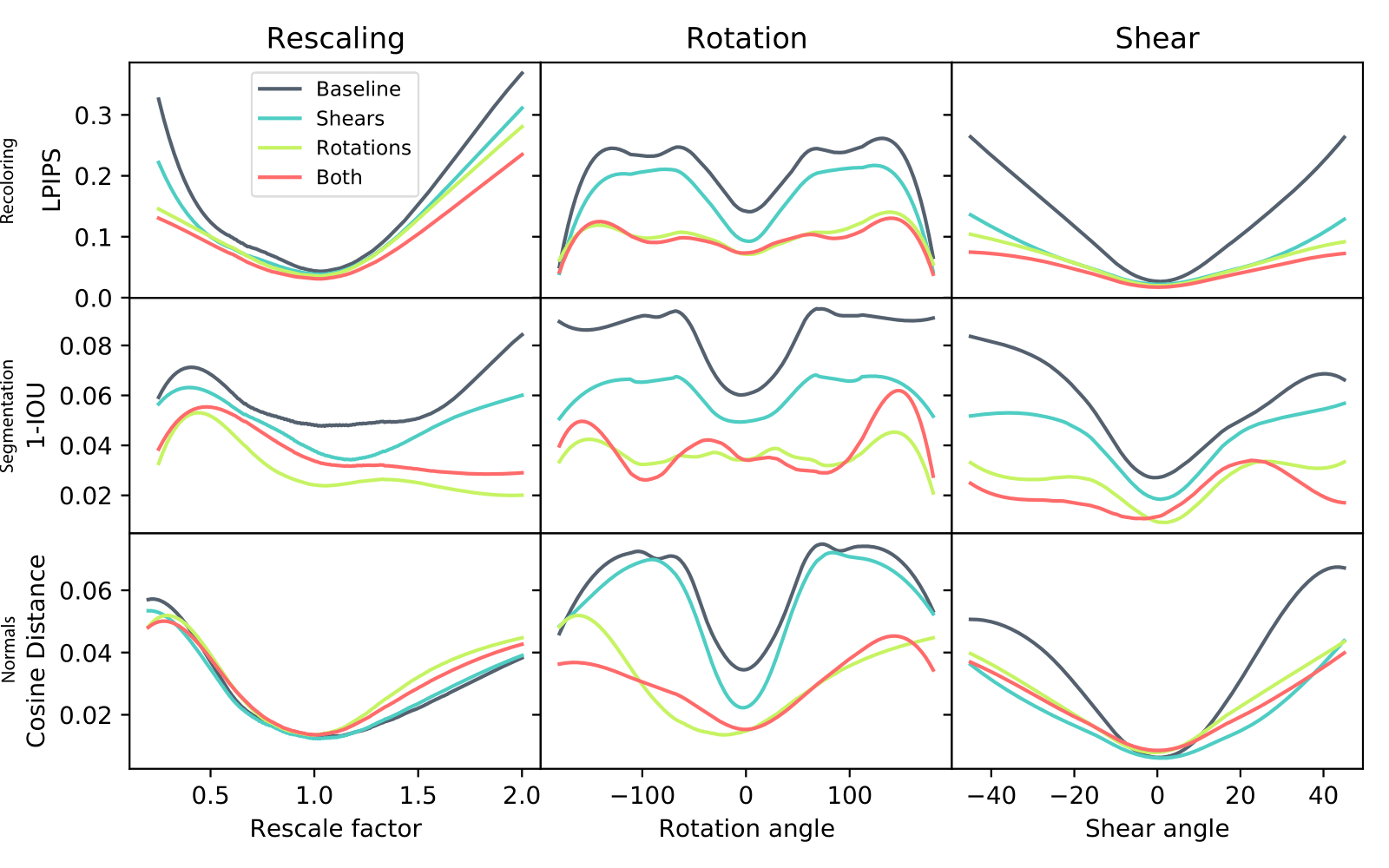}
	\vspace{-0.7cm}
	\caption{Quantitative evaluation of the robustness of our networks with respect to different transformations $\mathcal{T}$ (rescaling, rotation and shearing) of networks trained under different data augmentation policies for the \textit{denim} material. From top to bottom, we show results on: recoloring, segmentation, and normals estimation. Lower is better for each metric. }. 
	\vspace{-0.6cm}
	\label{fig:geometryinvariance}
\end{figure}

We measure their robustness with respect to three different transformations $\mathcal{T}$: rescalings, rotations and shears. As before, we use a guidance image taken under diffuse lighting as input for these experiments and attribute-specific distance metrics. 
Figure~\ref{fig:experiment1}~(b) shows that not adding those transforms generates visual artifacts for rotations and shears. Interestingly, without affine augmentations, the models hallucinates vertical yarns, as it is the only type of data that it has seen as input.
Figure~\ref{fig:geometryinvariance} shows quantitative metrics for the range of transformations $\mathcal{T}$ in which we evaluated the models.
Augmenting the training dataset with both shears and rotations generally provides the best results. Furthermore, this enhanced data augmentation policy improves the robustness of the models with respect to the scale of its inputs. Notably, applying random rotations during training appears to have a larger impact than random shears on the robustness of the models. 
This finding suggests that applying blind policies of data augmentation~\cite{antoniou2017data, shorten2019survey, sandfort2019data, chen2020simple} may not be an optimal strategy for some applications like ours, as the networks may not learn relevant features, or overfit to noise present in the training dataset.
Finally, it is worth noting that every network has the same number of parameters, and are trained for the same number of iterations, which shows that the generalization capabilities of the models can be increased at no extra training cost.

\section{Results and Comparisons}\label{sec:results}
\begin{figure}[t!]
	\includegraphics[width=0.9\linewidth]{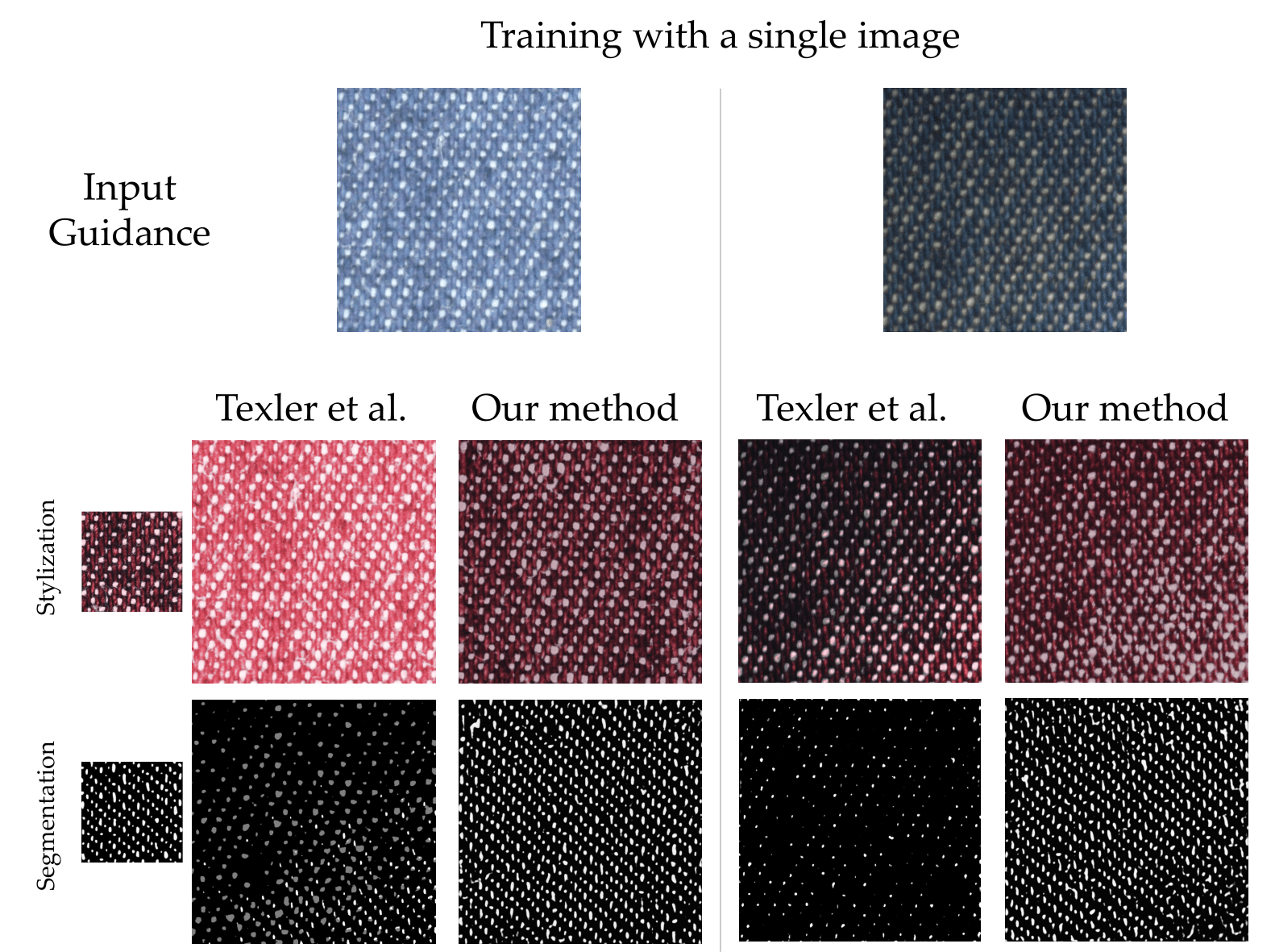}
	\vspace{-0.2cm}
	\caption{Comparison of our method with Texler~\etal~\cite{Texler20SIG} using a single diffuse image of the \emph{denim} material as training data.
		The task is to transfer the two attributes shown (\textit{Stylization} and \textit{Segmentation}) to two different guidance images.  
Even using a single image instead of a photometric dataset, we achieve higher quality mappings at a lower cost. Further results are included in the supplementary material. 
	}
	\label{fig:texler_comparison}
	\vspace{-0.5cm}
\end{figure}

In this section, we first compare our method with related approaches on image stylization and large scale SVBRDF material transfer. Then, we show the capabilities of our method to generalize to similar materials to those in their training set and present its limitations\footnote{In addition to the content presented in this manuscript and its supplementary material, we provide a \href{http://carlosrodriguezpardo.es/projects/NeuralPhotometricTransfer/}{web project} which contains further results and visualizations.}.
\vspace{-.2cm}

\subsection{Interactive Stylizations}
In the first category, the method of Texler~\etal~\cite{texler2020arbitrary} allows artists to interactively edit a few keyframes of a video and propagate that edition to the rest of the video. As ours, they formulate this transfer problem by training an encoder-decoder network using patch-based learning. But, contrarily to us, they do not perform any data augmentation policy outside of random cropping.
For a fair comparison with such method, we compare two setups: 1) using a single image as training input, and 2) using the photometric dataset. Results are shown in Figures~\ref{fig:teaser} and~\ref{fig:texler_comparison}. As Texler's method is not scale invariant, in both cases the training data provided for their model has the same scale as the images used for testing. Our models have been trained with the full policy of data augmentation.
In the first setup (Figure~\ref{fig:texler_comparison}), we use the diffuse illumination and hence compare their output with our \textit{diffuseNet} output.
As shown, none of the methods provide high quality results but our model manages to provide closer estimations.
In the second setup (Figure~\ref{fig:teaser}), we train their model with the photometric input. 
The best results are obtained with our method. Even though extending the input data using photometric cues has an impact on the quality of Texler's results, the lack of a data augmentation policy makes the transfer fuzzier and noisier. Further, their combination of style, adversarial and pixel-wise losses fails to yield predictable mappings. 
These results confirm the importance of a comprehensive data-augmentation policy, such as the one we propose when using neural networks for image processing tasks of this kind. In addition, our model is trained in less time with a smaller computational footprint (1 minute vs 5 minutes).

\begin{figure}[tb!]
	\centering
		\vspace{-0.2cm}
	\includegraphics[width=\linewidth]{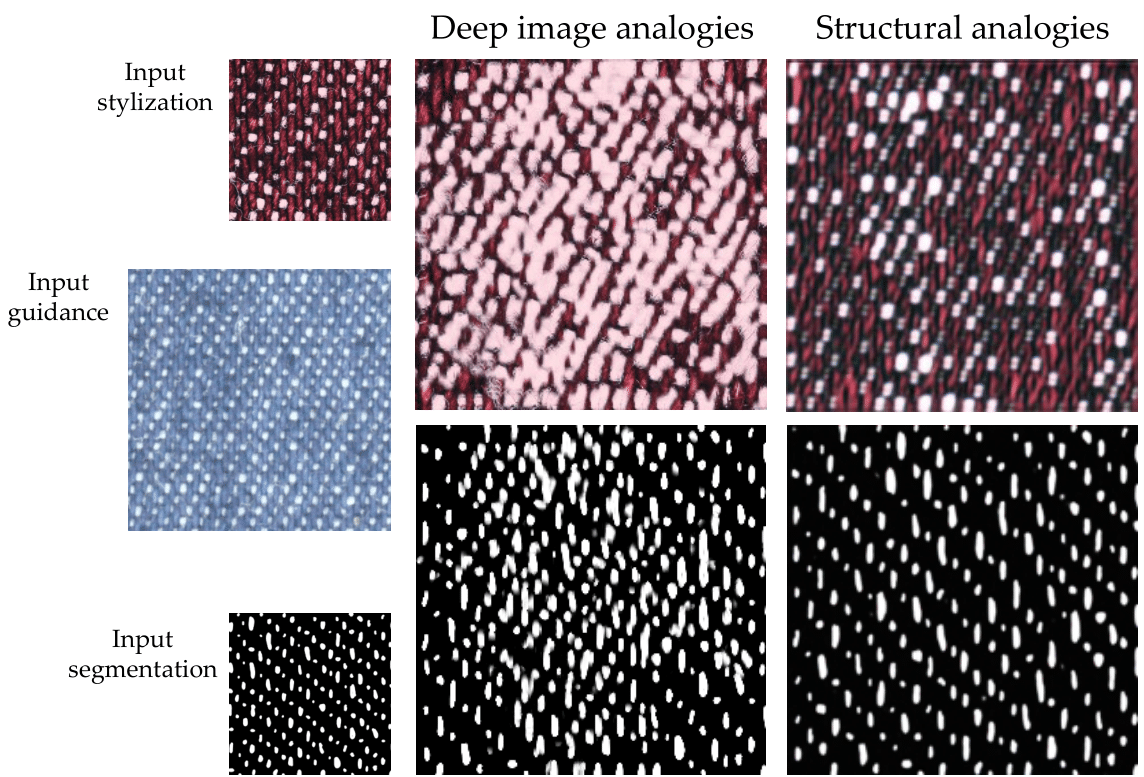}
	\vspace{-0.4cm}
	\caption{Comparison of image analogies approaches for input shown on the left, of the \emph{denim} material. From left to right: Deep Image Analogies~\cite{liao2017visual} and Structural Analogies~\cite{benaim2020structural}. Our results shows more accurate and predictable mappings, at less computational cost than the alternatives.
	}
	\label{fig:analogies_comparison}
	\vspace{-0.5cm}
\end{figure}

Another way of formulating this visual attribute transfer problem is through \textit{image analogies}. Using our single diffuse image for input, we compare our approach with two methods as shown in Figure~\ref{fig:analogies_comparison}. First, the work of Liao~\etal~\cite{liao2017visual}, that uses deep latent spaces as image descriptors; and the method of Benaim~\etal ~\cite{benaim2020structural} that trains single-image generative models to find bijective mappings between the \textit{structure} of one image and the \textit{style} of another.   
Our method qualitatively outperforms these methods with a fraction of the computational cost: 1 minute in our case, 10 minutes in~\cite{hertzmann2001image}, 40 minutes in~\cite{liao2017visual} and 10 hours in~\cite{benaim2020structural}. Once trained, our models can be used to evaluate any guidance image in real time for materials with similar microstructure. In contrast, image analogies methods require expensive optimizations for each guidance image. 
We refer the reader to the supplementary material for more comparisons with these methods.

\begin{figure}[tb!]
	\includegraphics[width=\linewidth]{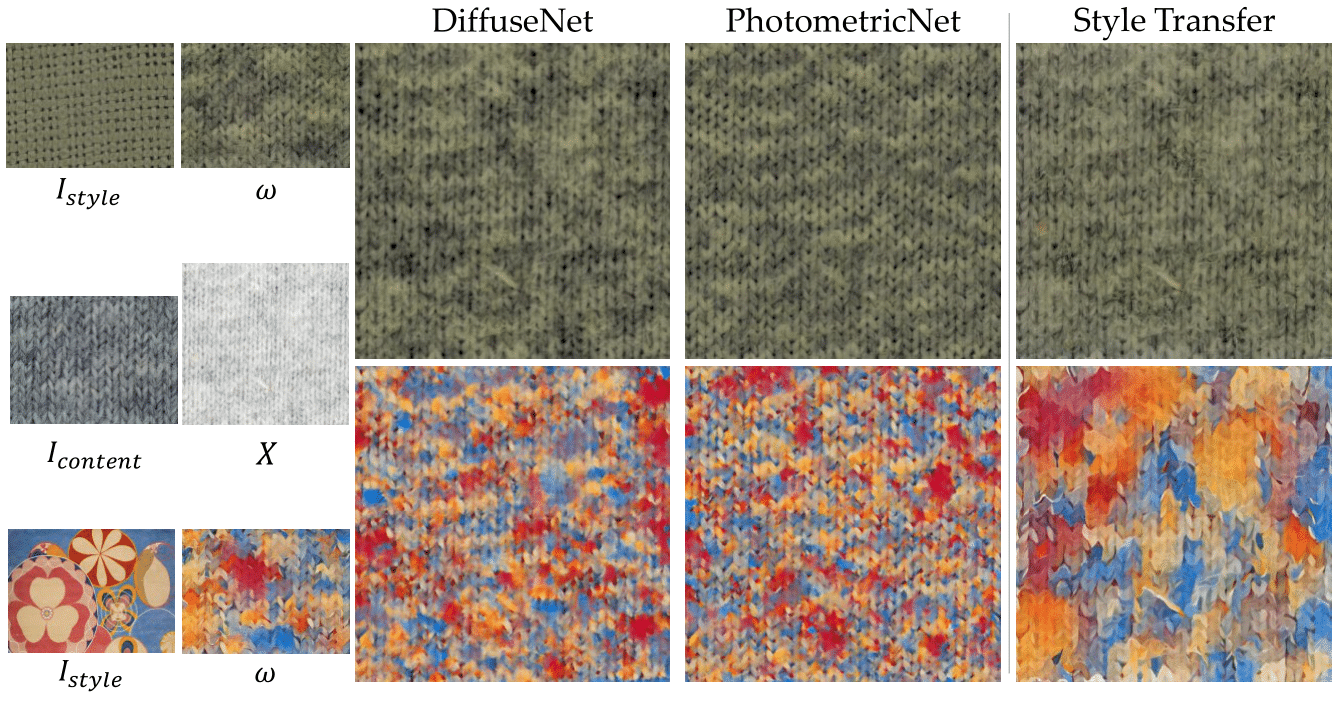}
	\vspace{-0.7cm}
	\caption{Our method allows for interactive material-aware visual attribute transfers. Using an off-the-shelf style transfer algorithm~\cite{gatys2015neural}, we can transfer the style of one image $I_{style}$ to the content of another $I_{content}$, obtaining a visual attribute $\omega$. Training a $\mathcal{M}_{\omega}$ to learn this relationship, we can find predictable style mappings, that we can transfer to guidance images $\imacro$, obtaining style transfers $\mathcal{M}_{\omega}(\imacro)$. Learning this transfer is inexpensive and allows for interactive editions. Performing this transfer directly to the guidance image generates artifacts and not-predictable mappings. 
	}
	\label{fig:style_transfer}
	\vspace{-0.3cm}
\end{figure}

Figure~\ref{fig:style_transfer} shows additional results of material stylizations. In these examples, we used the method of Gatys~\etal~\cite{gatys2015neural} to stylize a small patch of the material. Then, we trained a model using photometricNet and diffuseNet. As the guidance image we used a bigger image with diffuse illumination. Compared with naïve style transfer applied to the whole image, our approach provides detailed stylizations where the microstructure of the material is preserved. We further see that diffuseNet provides noisier results than photometricNet, probably due to the fact that the photometric cues help to preserve the local shading variations. We show more examples of this kind in the supplementary material.

\subsection{Creation of Large Scale Digital Material Assets}
\begin{figure*}[htb]
	\vspace{-0.1cm}
	\includegraphics[width=\linewidth]{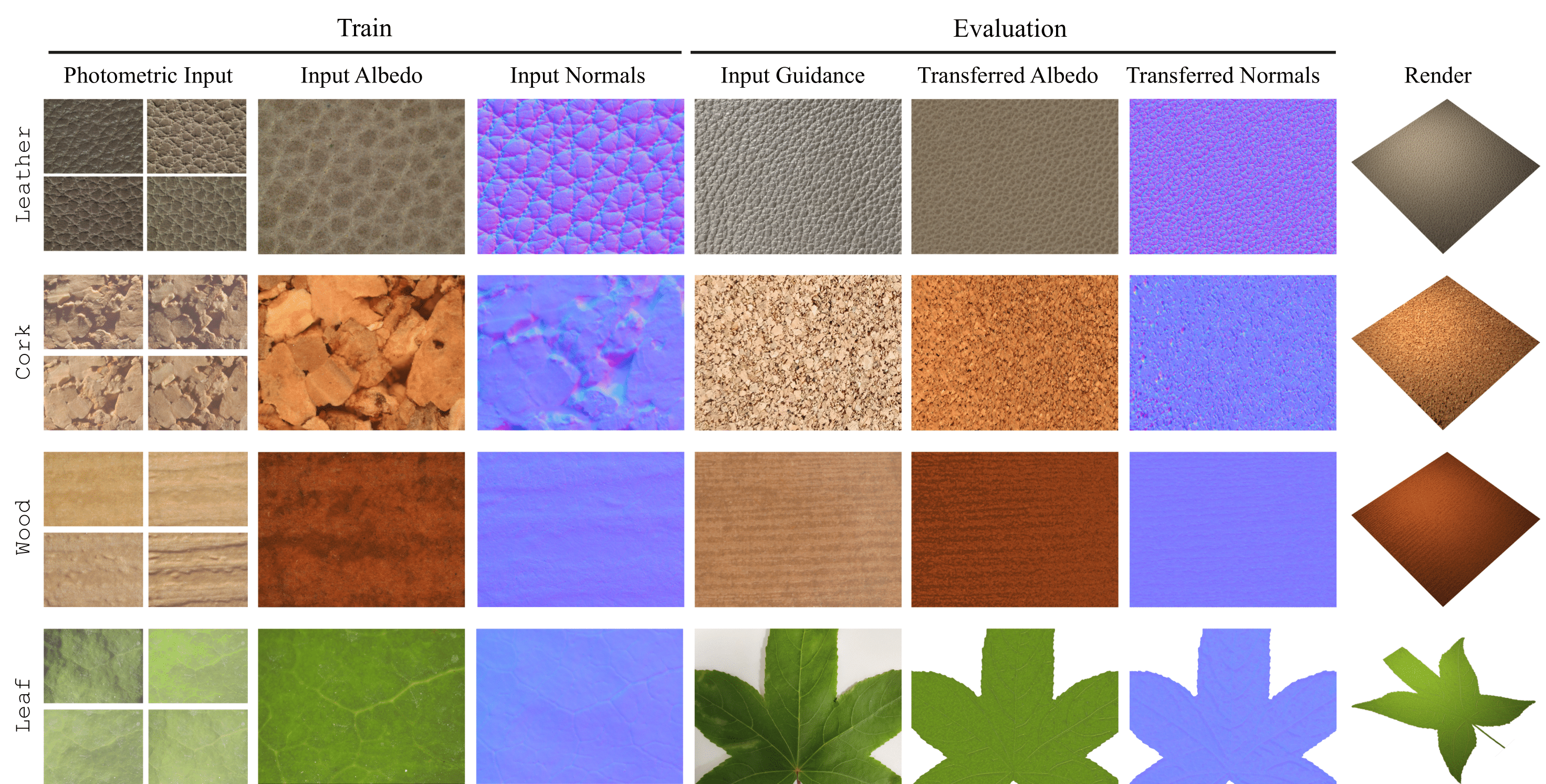}
	\vspace{-0.6cm}
	\caption{Results of our framework for material capture using a smartphone. Training two models, $\map_{a}$ and $\map_{n}$ with a photometric dataset, we obtain respectively albedo and normals from guidance images taken under uncontrolled conditions, which can be used by render engines. Here we have used Arnold~\cite{georgiev2018arnold} and a diffuse material model. Further examples are included in the supplementary material.
	}
	\vspace{-0.2cm}
	\label{fig:othermaterials}
\end{figure*}
\begin{figure*}[tb!]
	\centering
    \vspace{-0.3cm}
	\includegraphics[width=.92\linewidth]{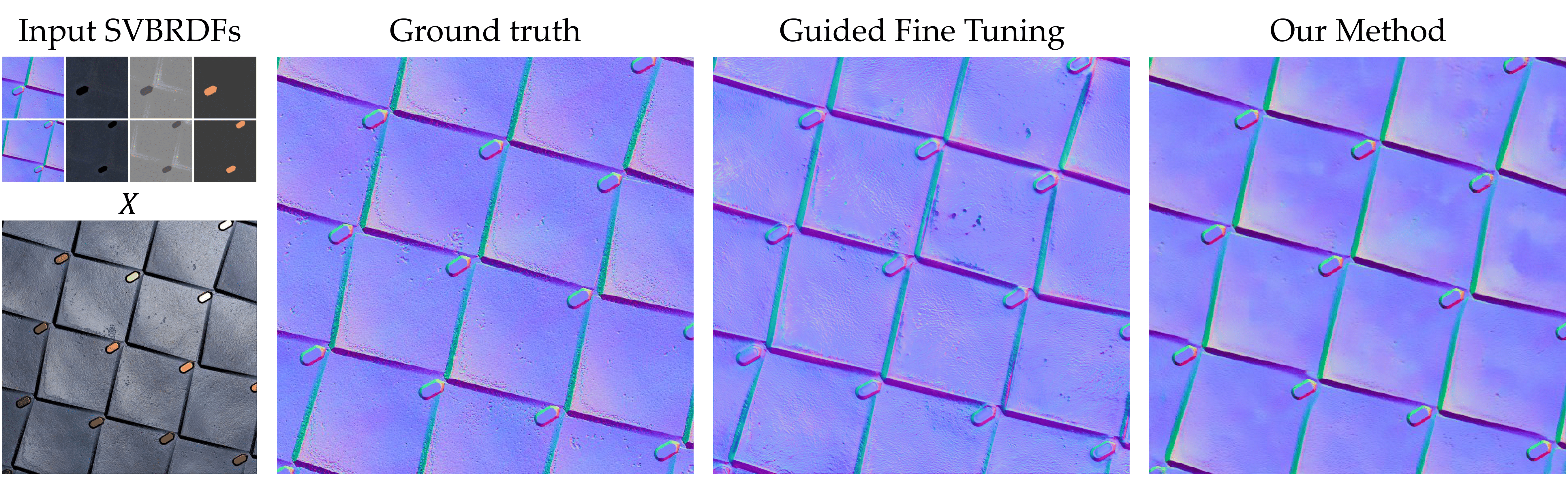}
    \vspace{-0.1cm}
	\caption{Comparison of our method with the Guided Fine Tuning, by Deschaintre~\etal~\cite{deschaintre2020guided}. Following their algorithm, we render the \textit{Input SVBRDFs}, and train a photometricNet on those renders. As shown, our method can achieve higher quality normal maps, with fewer artifacts. Input SVBRDFs, $\imacro$, ground truth and results from Guided Fine Tuning were obtained directly from~\cite{deschaintre2020guided}. Further examples are included in the supplementary material.}
	\label{fig:deschaintre}
	\vspace{-0.2cm}
\end{figure*}

\begin{figure*}[tb!]
	\centering
	\vspace{-0.2cm}
	\includegraphics[width=.92\linewidth]{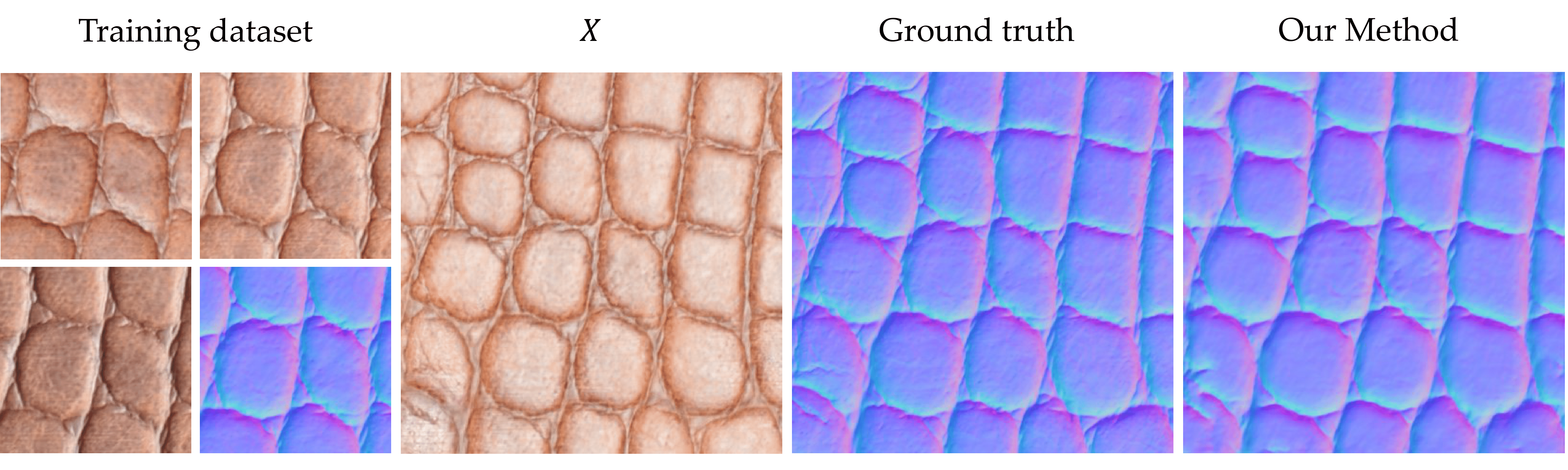}
		\vspace{-0.2cm}
	\caption{Our method can work with real BTF captured data. Training a photometricNet on a crop of the material (summarized in \textit{Training Dataset}), to output surface normals, we can estimate surface normals on larger areas of the material, even under novel viewing positions. }
	\label{fig:btf}
	\vspace{-0.2cm}
\end{figure*}

Our method can be used to propagate SVBRDFs estimated locally in a small area of the material to larger samples. Figure~\ref{fig:othermaterials} illustrates for a diverse set of materials that we can propagate albedo and normals estimated at high resolution in a small area of 0.7 x 0.9 cm to guidance images of 13 x 13 cm taken with a smartphone. Even though we train the albedo and normals models separately, which does not guarantee pixel-wise coherence between the estimated maps, the rendered images show realistic-looking materials, even with a diffuse material model. As opposed to the method of Deschaintre~\etal~\cite{deschaintre2020guided}, which is trained to output directly Cook-Torrance~\cite{cook1982reflectance} material layers, our method is agnostic to the parameters of the SVBRDF. 

We assess the capabilities of our method on this setting using the same SVBRDF propagation scenario as proposed in~\cite{deschaintre2020guided}. 
Using a small crop of a synthetic SVBRDF as input, we render 27 images using the same directional light position as we used in our real dataset, and train a photometricNet to estimate the surface normals from each of those renders. We then evaluate this model using a larger area of the material, illuminated under an unknown lighting position. 
In Table~\ref{tab:comparison_deschaintre}, we show a quantitative comparison with~\cite{deschaintre2020guided}, under different image quality metrics. As shown, our method achieves better scores on pixel-wise metrics, whilst~\cite{deschaintre2020guided} achieves better deep perceptual scores, as in the LPIPS metric~\cite{zhang2018unreasonable}. This might be related to the design of our loss function: we directly minimize pixel-wise differences, while~\cite{deschaintre2020guided} is optimized using a render-aware loss.    
Qualitatively, as shown on Figure~\ref{fig:deschaintre}, our method obtains comparable quality mappings, with fewer artifacts.
We refer the reader to the supplementary material for a larger pool of examples with a diverse set of materials. 
Despite capturing SVBRDF or arbitrary materials is not the goal of our method, we include in the supplementary comparisons with a direct SVBRDF acquisition method~\cite{gao2019deep} using our own real dataset. 

A similar setting can be used to propagate attributes leveraging BTF measurements as training data. Figure~\ref{fig:btf} shows an example using a BTF from~\cite{weinmann-2014-materialclassification}. Using Photometric Stereo~\cite{ikeuchi1981determining}, we compute their surface normals and train a photometricNet on a central crop of the BTF, using the captures at a camera position of $(\phi = 0\degree, \theta = 0\degree)$. We then evaluate this model using the full material surface, and a novel camera position, of $(\phi = 15\degree, \theta = 11\degree)$. As shown, our model is capable of working with captured BTF data. We provide further examples on the supplementary material.

\begin{table}[tb!]
	\centering
	\resizebox{\columnwidth}{!}{%
		\begin{tabular}{@{}ccccccccc@{}}
			\toprule
			\multirow{2}{*}{\textbf{Material ID}} & \multicolumn{2}{c}{\textbf{SSIM} $\uparrow$}  & \multicolumn{2}{c}{\textbf{PSNR} $\uparrow$} & \multicolumn{2}{c}{\textbf{MSE} $\downarrow$}& \multicolumn{2}{c}{\textbf{LPIPS} $\downarrow$} \\ \cmidrule(l){2-9} 
			& \cite{deschaintre2020guided} & \textbf{Ours}             & \cite{deschaintre2020guided} & \textbf{Ours}              & \cite{deschaintre2020guided} & \textbf{Ours}             & \cite{deschaintre2020guided} & \textbf{Ours}        \\ \midrule
			\multicolumn{1}{c|}{\textbf{560}}  & 0,905                        & 0,925                     & 31,190                       & 32,310                     & 0,002                        & 0,002                     & 0,244                        & 0,221                \\
			\multicolumn{1}{c|}{\textbf{1581}} & 0,645                        & 0,675                     & 29,740                       & 29,920                     & 0,006                        & 0,005                     & 0,398                        & 0,446                \\
			\multicolumn{1}{c|}{\textbf{1684}} & 0,654                        & 0,746                     & 29,711                       & 30,410                     & 0,007                        & 0,003                     & 0,196                        & 0,251                \\
			\multicolumn{1}{c|}{\textbf{2111}} & 0,770                        & 0,783                     & 32,250                       & 32,881                     & 0,002                        & 0,002                     & 0,397                        & 0,383                \\
			\multicolumn{1}{c|}{\textbf{Average}}                  & 0,744                        & \textbf{0,782} & 30,723                       & \textbf{31,380} & 0,004                        & \textbf{0,003} & \textbf{0,309}                        & 0,325                \\ \bottomrule
		\end{tabular}%
	}
	\caption{Quantitative comparison with~\cite{deschaintre2020guided}, on the studied materials and different performance metrics. As shown, our method provides better pixel-wise accuracy than~\cite{deschaintre2020guided}, while their method obtains better perceptual scores.}\vspace{-.5cm}
	\label{tab:comparison_deschaintre}
\end{table}

\subsection{Generalization to Similar Materials}

\begin{figure}[tb!]
	\vspace{-0.0cm}
	\centering
	\includegraphics[width=\linewidth]{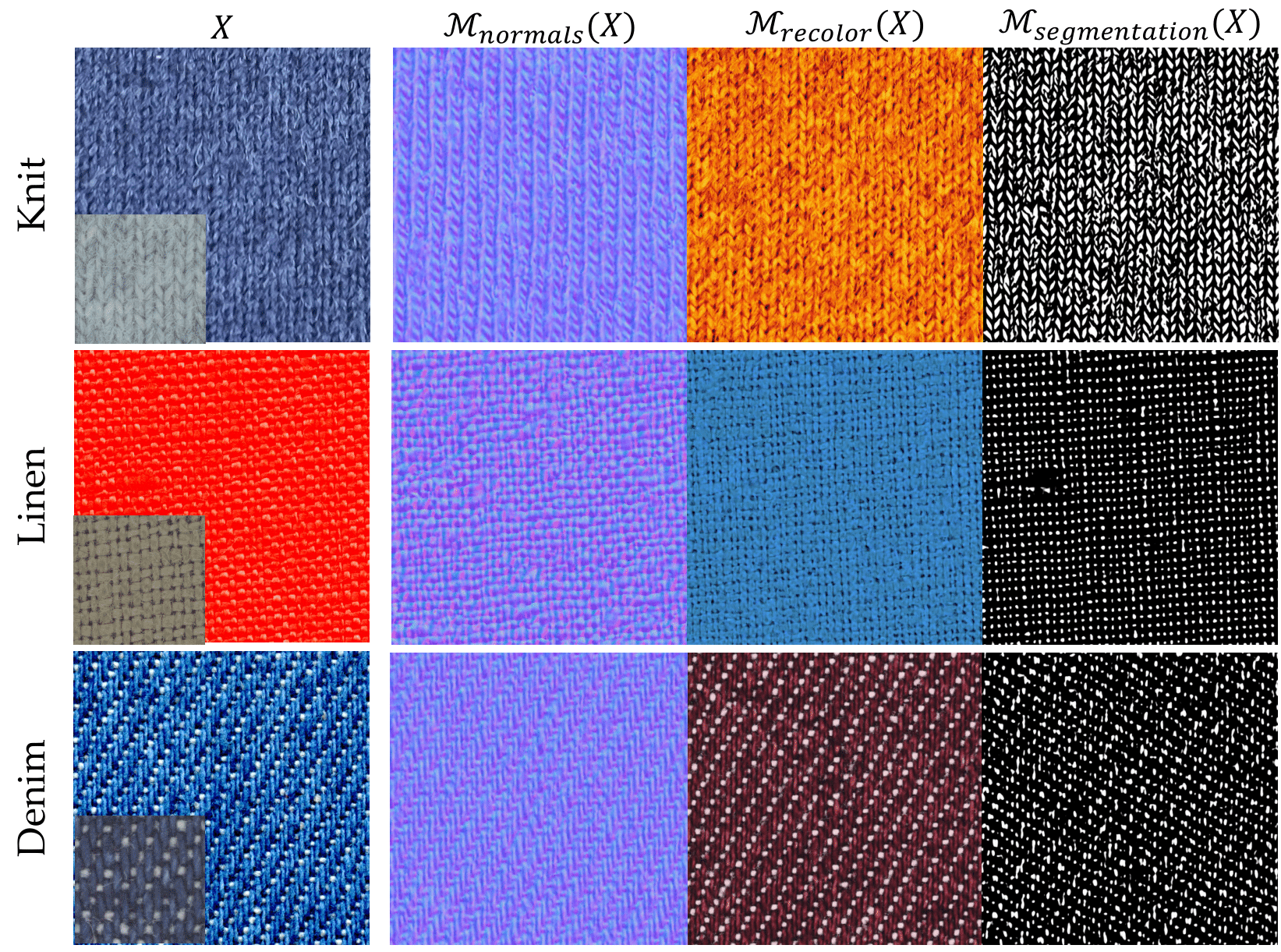}
	\vspace{-.5cm}
	\caption{Generalization capabilities of our method when evaluated on materials similar to those in their training dataset. On the top row, we show the outputs of the model trained on the \emph{knit} in our dataset (Figure~\ref{fig:dataset}), and evaluated on a different guidance images.  We also show examples on \emph{linen} and \emph{denim} fabrics, with different conditions of saturation, blur, scale and illumination. Even in very challenging cases, where the structure of the material is barely visible, as in the overly-saturated red linen or the noisy blue knit fabrics, our photometricNets can yield plausible results. We include further results and the training datasets in the supplementary material. The insets represent the training dataset by each model, as represented in Figure~\ref{fig:dataset}.}
	\label{fig:generalization_similar_materials2}
	\vspace{-0.5cm}
\end{figure}

In previous experiments, we have shown the performance of our models when the guidance image corresponds to the material used for training. 
In this experiment, we show that our models, despite being trained only on a set of captures of a single material, generalize to materials of the same category.
Figure~\ref{fig:generalization_similar_materials2} shows some examples for models trained using our dataset (Figure~\ref{fig:dataset}), taking input guidance images of different albedos and scales. The transfer works thanks to the network design and data augmentation strategy that is designed to use microstructure details as guidance.
Our method could thus be used to transfer visual attributes for a diverse set of materials by simply training one model using a single but representative material of each category.

\subsection{Limitations}\label{sec:limitations}
\begin{figure}[tb!]
	\centering
	\includegraphics[width=\linewidth]{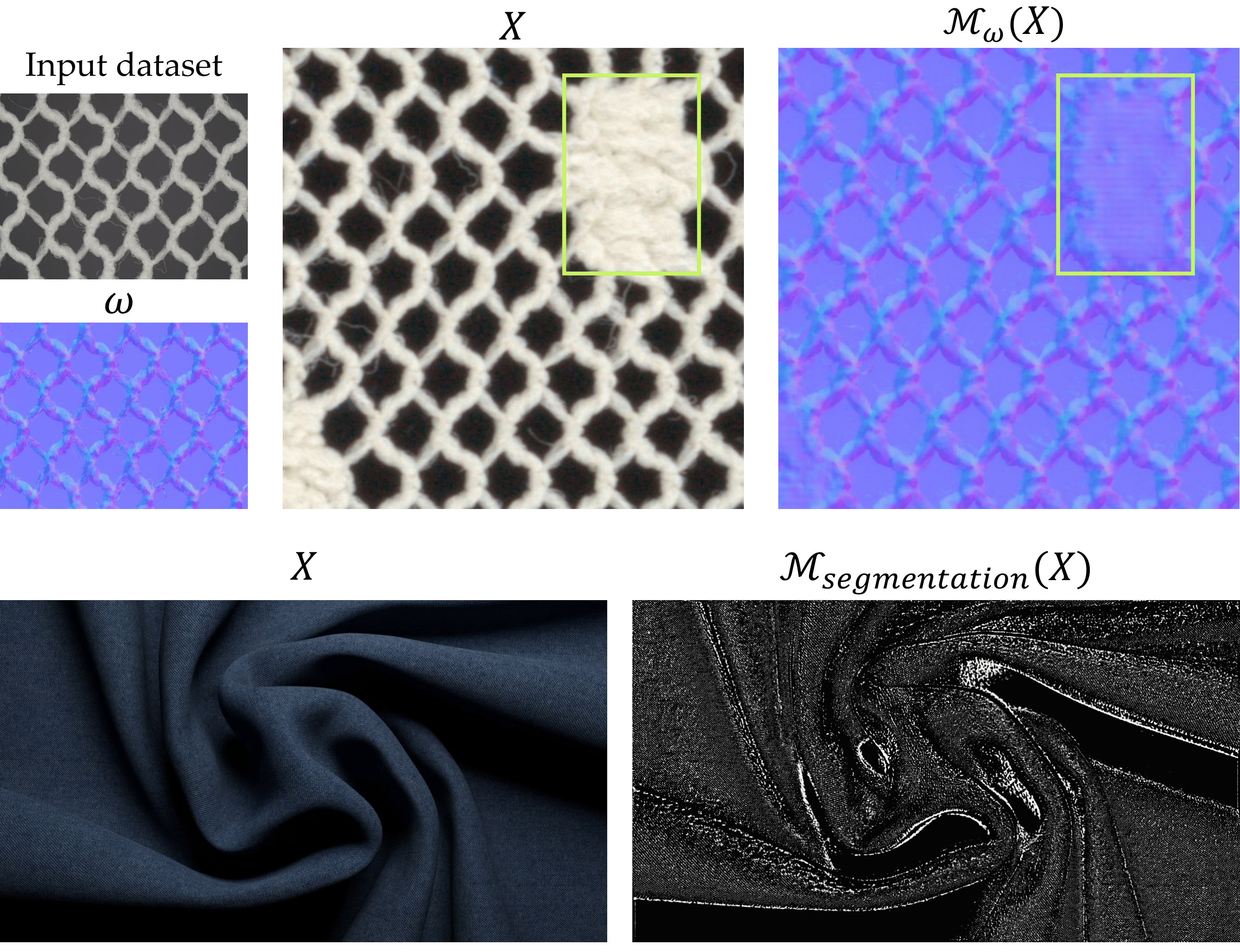}
	\vspace{-.6cm}
	\caption{Failure cases of our method. As shown on the first row, if the input dataset does not represent the heterogeneity present in the guidance $\imacro$, the model fails to yield compelling results on unseen structures of the material, as shown on the green box. On the second row, we show a guidance image of the \emph{denim} material $\imacro$ which exhibits strong geometric and illumination variations, outside of the range in which we train $\mathcal{M}$ with. As such, the model shows a poor performance on the segmentation task. }
	\label{fig:failure_case}
	\vspace{-0.4cm}
\end{figure}

Our models are not guaranteed to provide high-quality results outside the range of input data and data augmentation policies we train them on. This limitation is common to all learning-based approaches. It is unlikely that the framework is capable of generalizing to resolutions higher to those of the training data; or down-sampled images in which the texture details are not recognizable. Furthermore, the type of transformations we apply to the training data may not represent all the possible geometric variations, or non-linear warpings that materials are subject to in the real world. For materials which exhibit a strong variation in their microstructure, which cannot be fully captured using a single photometric dataset, our patch-based approach will likely fail to generalize to the full heterogeneity. Figure~\ref{fig:failure_case} shows two examples of failure cases for which the test data is not included in the training data and where the affine and illumination transformations are outside the suitable range. Further examples of these limitations are included in the supplementary material.

\section{Conclusions and Future Work}

In this paper, we have proposed a neural visual attribute transfer framework capable of transferring, for a given material, many types of visual property maps to images of unseen patches of the same -or similar- material taken under different illumination, capture setup, and affine distortion. 
To our knowledge, the proposed framework is the first method capable of leveraging the optical behavior of the material to this purpose by being trained using a photometric approach. 
Such an approach, besides the illumination-invariance we have shown, helps the neural network learn better mappings between visual domains, finding a physically-based representation of the material. 
Further, we have presented a comprehensive policy of data augmentation which outperforms previous work on visual attribute transfer given a single image of the material.

We have shown that our method can be used to transfer any kind of visual attribute estimated locally to larger material samples. Further, we have demonstrated that our models, although trained on a single material, generalize to materials of the same category. We think our findings will inspire future work showing that smart training strategies might alleviate the need for massive datasets.

Our method could be extended in several ways. The need for obtaining high-resolution captures taken under different illuminations may be reduced by generating rendering images through recent advances in inverse material acquisition~\cite{dong2019deep}. Further training with multiple patches may help to cope with material heterogeneity~\cite{Texler20SIG}. Similarly, our findings suggest that extending the data augmentation policy to include $3D$ deformations will likely improve the accuracy.
Beyond the generation of large scale digital assets for rendering, our method may have potential in other visual computing applications that require a low level understanding of the properties of the materials in real scenes. For example, the yarn segmentation application shown in the paper might be suitable as input to shape from texture applications. Specific visual attributes might be useful to identify or highlight defects for image forensics problems; or to enhance different features in real-time or AR applications. 

\paragraph*{\textbf{Acknowledgements}}
Elena Garces was partially supported by a Torres Quevedo Fellowship (PTQ2018-009868). We thank Jorge López-Moreno for his feedback and David Pascual for his help with the creation of property maps.

\bibliographystyle{IEEEtran}
\bibliography{references.bib}   

\begin{thebibliography}{10}
\providecommand{\url}[1]{#1}
\csname url@samestyle\endcsname
\providecommand{\newblock}{\relax}
\providecommand{\bibinfo}[2]{#2}
\providecommand{\BIBentrySTDinterwordspacing}{\spaceskip=0pt\relax}
\providecommand{\BIBentryALTinterwordstretchfactor}{4}
\providecommand{\BIBentryALTinterwordspacing}{\spaceskip=\fontdimen2\font plus
\BIBentryALTinterwordstretchfactor\fontdimen3\font minus
  \fontdimen4\font\relax}
\providecommand{\BIBforeignlanguage}[2]{{%
\expandafter\ifx\csname l@#1\endcsname\relax
\typeout{** WARNING: IEEEtran.bst: No hyphenation pattern has been}%
\typeout{** loaded for the language `#1'. Using the pattern for}%
\typeout{** the default language instead.}%
\else
\language=\csname l@#1\endcsname
\fi
#2}}
\providecommand{\BIBdecl}{\relax}
\BIBdecl

\bibitem{steinhausen2014acquiring}
H.~C. Steinhausen, D.~den Brok, M.~B. Hullin, and R.~Klein, ``Acquiring
  bidirectional texture functions for large-scale material samples,'' 2014.

\bibitem{Texler20SIG}
O.~Texler, D.~Futschik, M.~Ku\v{c}era, O.~Jamri\v{s}ka, \v{S}\'{a}rka
  Sochorov\'{a}, M.~Chai, S.~Tulyakov, and D.~S\'{y}kora, ``Interactive video
  stylization using few-shot patch-based training,'' \emph{ACM Transactions on
  Graphics}, vol.~39, no.~4, p.~73, 2020.

\bibitem{hertzmann2001image}
A.~Hertzmann, C.~E. Jacobs, N.~Oliver, B.~Curless, and D.~H. Salesin, ``Image
  analogies,'' in \emph{Proceedings of the 28th annual conference on Computer
  graphics and interactive techniques}, 2001, pp. 327--340.

\bibitem{melendez2012transfer}
F.~Melendez, M.~Glencross, J.~Starck, and G.~J. Ward, ``Transfer of albedo and
  local depth variation to photo-textures,'' in \emph{Proceedings of the 9th
  European Conference on Visual Media Production}, 2012, pp. 40--48.

\bibitem{riviere2016mobile}
J.~Riviere, P.~Peers, and A.~Ghosh, ``Mobile surface reflectometry,'' in
  \emph{Computer Graphics Forum}, vol.~35, no.~1.\hskip 1em plus 0.5em minus
  0.4em\relax Wiley Online Library, 2016, pp. 191--202.

\bibitem{mazlov2019neural}
I.~Mazlov, S.~Merzbach, E.~Trunz, and R.~Klein, ``Neural appearance synthesis
  and transfer,'' 2019.

\bibitem{benard2013stylizing}
P.~B{\'e}nard, F.~Cole, M.~Kass, I.~Mordatch, J.~Hegarty, M.~S. Senn,
  K.~Fleischer, D.~Pesare, and K.~Breeden, ``Stylizing animation by example,''
  \emph{ACM Transactions on Graphics (TOG)}, vol.~32, no.~4, pp. 1--12, 2013.

\bibitem{barnes2015patchtable}
C.~Barnes, F.-L. Zhang, L.~Lou, X.~Wu, and S.-M. Hu, ``Patchtable: Efficient
  patch queries for large datasets and applications,'' \emph{ACM Transactions
  on Graphics (ToG)}, vol.~34, no.~4, pp. 1--10, 2015.

\bibitem{jamrivska2019stylizing}
O.~Jamri{\v{s}}ka, {\v{S}}.~Sochorov{\'a}, O.~Texler, M.~Luk{\'a}{\v{c}},
  J.~Fi{\v{s}}er, J.~Lu, E.~Shechtman, and D.~S{\`y}kora, ``Stylizing video by
  example,'' \emph{ACM Transactions on Graphics (TOG)}, vol.~38, no.~4, pp.
  1--11, 2019.

\bibitem{reed2015deep}
S.~E. Reed, Y.~Zhang, Y.~Zhang, and H.~Lee, ``Deep visual analogy-making,'' in
  \emph{Advances in Neural Information Processing Systems}, 2015, pp.
  1252--1260.

\bibitem{gatys2015neural}
L.~A. Gatys, A.~S. Ecker, and M.~Bethge, ``A neural algorithm of artistic
  style,'' \emph{arXiv preprint arXiv:1508.06576}, 2015.

\bibitem{simonyan2014very}
K.~Simonyan and A.~Zisserman, ``Very deep convolutional networks for
  large-scale image recognition,'' in \emph{3rd International Conference on
  Learning Representations, {ICLR}}, 2015.

\bibitem{deng2009imagenet}
J.~Deng, W.~Dong, R.~Socher, L.-J. Li, K.~Li, and L.~Fei-Fei, ``Imagenet: A
  large-scale hierarchical image database,'' in \emph{Proceedings of the IEEE
  Conference on Computer Vision and Pattern Recognition}, 2009, pp. 248--255.

\bibitem{huang2017arbitrary}
X.~Huang and S.~Belongie, ``Arbitrary style transfer in real-time with adaptive
  instance normalization,'' in \emph{Proceedings of the IEEE International
  Conference on Computer Vision}, 2017, pp. 1501--1510.

\bibitem{li2017universal}
Y.~Li, C.~Fang, J.~Yang, Z.~Wang, X.~Lu, and M.-H. Yang, ``Universal style
  transfer via feature transforms,'' in \emph{Advances in Neural Information
  Processing Systems}, 2017, pp. 386--396.

\bibitem{johnson2016perceptual}
J.~Johnson, A.~Alahi, and L.~Fei-Fei, ``Perceptual losses for real-time style
  transfer and super-resolution,'' in \emph{Proceedings of the European
  Conference on Computer vision (ECCV)}, 2016, pp. 694--711.

\bibitem{chen2017stylebank}
D.~Chen, L.~Yuan, J.~Liao, N.~Yu, and G.~Hua, ``Stylebank: An explicit
  representation for neural image style transfer,'' in \emph{Proceedings of the
  IEEE Conference on Computer Vision and Pattern Recognition}, 2017, pp.
  1897--1906.

\bibitem{chen2017coherent}
D.~Chen, J.~Liao, L.~Yuan, N.~Yu, and G.~Hua, ``Coherent online video style
  transfer,'' in \emph{Proceedings of the IEEE International Conference on
  Computer Vision}, 2017, pp. 1105--1114.

\bibitem{zhang2018unreasonable}
R.~Zhang, P.~Isola, A.~A. Efros, E.~Shechtman, and O.~Wang, ``The unreasonable
  effectiveness of deep features as a perceptual metric,'' in \emph{Proceedings
  of the IEEE Conference on Computer Vision and Pattern Recognition}, 2018, pp.
  586--595.

\bibitem{gatys2017controlling}
L.~A. Gatys, A.~S. Ecker, M.~Bethge, A.~Hertzmann, and E.~Shechtman,
  ``Controlling perceptual factors in neural style transfer,'' in
  \emph{Proceedings of the IEEE Conference on Computer Vision and Pattern
  Recognition}, 2017, pp. 3985--3993.

\bibitem{gu2018arbitrary}
S.~Gu, C.~Chen, J.~Liao, and L.~Yuan, ``Arbitrary style transfer with deep
  feature reshuffle,'' in \emph{Proceedings of the IEEE Conference on Computer
  Vision and Pattern Recognition}, 2018, pp. 8222--8231.

\bibitem{jing2019neural}
Y.~Jing, Y.~Yang, Z.~Feng, J.~Ye, Y.~Yu, and M.~Song, ``Neural style transfer:
  A review,'' \emph{IEEE Transactions on Visualization and Computer Graphics},
  2019.

\bibitem{liao2017visual}
J.~Liao, Y.~Yao, L.~Yuan, G.~Hua, and S.~B. Kang, ``Visual attribute transfer
  through deep image analogy,'' \emph{ACM Transactions on Graphics (TOG)},
  vol.~36, no.~4, pp. 1--15, 2017.

\bibitem{shaham2019singan}
T.~R. Shaham, T.~Dekel, and T.~Michaeli, ``Singan: Learning a generative model
  from a single natural image,'' in \emph{Proceedings of the IEEE International
  Conference on Computer Vision}, 2019, pp. 4570--4580.

\bibitem{benaim2020structural}
S.~Benaim, R.~Mokady, A.~Bermano, and L.~Wolf, ``Structural analogy from a
  single image pair,'' in \emph{Computer Graphics Forum}.\hskip 1em plus 0.5em
  minus 0.4em\relax Wiley Online Library, 2020.

\bibitem{fivser2016stylit}
J.~Fi{\v{s}}er, O.~Jamri{\v{s}}ka, M.~Luk{\'a}{\v{c}}, E.~Shechtman, P.~Asente,
  J.~Lu, and D.~S{\`y}kora, ``Stylit: illumination-guided example-based
  stylization of 3d renderings,'' \emph{ACM Transactions on Graphics (TOG)},
  vol.~35, no.~4, pp. 1--11, 2016.

\bibitem{he2018deep}
M.~He, D.~Chen, J.~Liao, P.~V. Sander, and L.~Yuan, ``Deep exemplar-based
  colorization,'' \emph{ACM Transactions on Graphics (TOG)}, vol.~37, no.~4,
  pp. 1--16, 2018.

\bibitem{zhang2019deep}
B.~Zhang, M.~He, J.~Liao, P.~V. Sander, L.~Yuan, A.~Bermak, and D.~Chen, ``Deep
  exemplar-based video colorization,'' in \emph{Proceedings of the IEEE
  Conference on Computer Vision and Pattern Recognition}, 2019, pp. 8052--8061.

\bibitem{he2019progressive}
M.~He, J.~Liao, D.~Chen, L.~Yuan, and P.~V. Sander, ``Progressive color
  transfer with dense semantic correspondences,'' \emph{ACM Transactions on
  Graphics (TOG)}, vol.~38, no.~2, pp. 1--18, 2019.

\bibitem{an2008appprop}
X.~An and F.~Pellacini, ``Appprop: all-pairs appearance-space edit
  propagation,'' \emph{ACM Transactions on Graphics (TOG)}, vol.~27, no.~3, pp.
  1--9, 2008.

\bibitem{endo2016deepprop}
Y.~Endo, S.~Iizuka, Y.~Kanamori, and J.~Mitani, ``Deepprop: Extracting deep
  features from a single image for edit propagation,'' in \emph{Computer
  Graphics Forum}, vol.~35, no.~2.\hskip 1em plus 0.5em minus 0.4em\relax Wiley
  Online Library, 2016, pp. 189--201.

\bibitem{li2008scribbleboost}
Y.~Li, E.~Adelson, and A.~Agarwala, ``Scribbleboost: Adding classification to
  edge-aware interpolation of local image and video adjustments,'' in
  \emph{Computer Graphics Forum}, vol.~27, no.~4.\hskip 1em plus 0.5em minus
  0.4em\relax Wiley Online Library, 2008, pp. 1255--1264.

\bibitem{dana1999reflectance}
K.~J. Dana, B.~Van~Ginneken, S.~K. Nayar, and J.~J. Koenderink, ``Reflectance
  and texture of real-world surfaces,'' \emph{ACM Transactions On Graphics
  (TOG)}, vol.~18, no.~1, pp. 1--34, 1999.

\bibitem{leung2001representing}
T.~Leung and J.~Malik, ``Representing and recognizing the visual appearance of
  materials using three-dimensional textons,'' \emph{International Journal of
  Computer Vision}, vol.~43, no.~1, pp. 29--44, 2001.

\bibitem{rainer2019neural}
G.~Rainer, W.~Jakob, A.~Ghosh, and T.~Weyrich, ``Neural btf compression and
  interpolation,'' in \emph{Computer Graphics Forum}, vol.~38, no.~2.\hskip 1em
  plus 0.5em minus 0.4em\relax Wiley Online Library, 2019, pp. 235--244.

\bibitem{rainer2020unified}
G.~Rainer, A.~Ghosh, W.~Jakob, and T.~Weyrich, ``Unified neural encoding of
  btfs,'' in \emph{Computer Graphics Forum}, vol.~39, no.~2.\hskip 1em plus
  0.5em minus 0.4em\relax Eurographics Association, 2020, pp. 1--13.

\bibitem{steinhausen2015crossdevice}
H.~C. Steinhausen, D.~den Brok, M.~B. Hullin, and R.~Klein, ``Extrapolating
  large-scale material btfs under cross-device constraints,'' in \emph{Vision,
  Modeling {\&} Visualization}, D.~Bommes, T.~Ritschel, and T.~Schultz,
  Eds.\hskip 1em plus 0.5em minus 0.4em\relax The Eurographics Association,
  2015, pp. 143--150.

\bibitem{steinhausen2015normals}
H.~C. Steinhausen, R.~Mart{\'i}n, D.~den Brok, M.~B. Hullin, and R.~Klein,
  ``Extrapolation of bidirectional texture functions using texture synthesis
  guided by photometric normals,'' in \emph{Measuring, Modeling, and
  Reproducing Material Appearance II (SPIE 9398)}, vol. 9398, no.~14, Feb.
  2015.

\bibitem{diamanti2015synthesis}
O.~Diamanti, C.~Barnes, S.~Paris, E.~Shechtman, and O.~Sorkine-Hornung,
  ``Synthesis of complex image appearance from limited exemplars,'' \emph{ACM
  Transactions on Graphics (TOG)}, vol.~34, no.~2, pp. 1--14, 2015.

\bibitem{aittala2015two}
M.~Aittala, T.~Weyrich, and J.~Lehtinen, ``Two-shot svbrdf capture for
  stationary materials,'' \emph{ACM Transactions on Graphics (TOG)}, vol.~34,
  no.~4, pp. 1--13, 2015.

\bibitem{guehl2020semi}
P.~Guehl, R.~All{\`e}gre, J.-M. Dischler, B.~Benes, and E.~Galin,
  ``Semi-procedural textures using point process texture basis functions,'' in
  \emph{Computer Graphics Forum}, vol.~39, no.~4.\hskip 1em plus 0.5em minus
  0.4em\relax Wiley Online Library, 2020, pp. 159--171.

\bibitem{lefebvre2006appearance}
S.~Lefebvre and H.~Hoppe, ``Appearance-space texture synthesis,'' \emph{ACM
  Transactions on Graphics (TOG)}, vol.~25, no.~3, pp. 541--548, 2006.

\bibitem{elad2017style}
M.~Elad and P.~Milanfar, ``Style transfer via texture synthesis,'' \emph{IEEE
  Transactions on Image Processing}, vol.~26, no.~5, pp. 2338--2351, 2017.

\bibitem{zhou2018non}
Y.~Zhou, Z.~Zhu, X.~Bai, D.~Lischinski, D.~Cohen-Or, and H.~Huang,
  ``Non-stationary texture synthesis by adversarial expansion,'' \emph{ACM
  Transactions on Graphics (TOG)}, vol.~37, no.~4, pp. 1--13, 2018.

\bibitem{fruhstuck2019tilegan}
A.~Fr{\"u}hst{\"u}ck, I.~Alhashim, and P.~Wonka, ``Tilegan: synthesis of
  large-scale non-homogeneous textures,'' \emph{ACM Transactions on Graphics
  (TOG)}, vol.~38, no.~4, pp. 1--11, 2019.

\bibitem{rodriguez2019automatic}
C.~Rodriguez-Pardo, S.~Suja, D.~Pascual, J.~Lopez-Moreno, and E.~Garces,
  ``Automatic extraction and synthesis of regular repeatable patterns,''
  \emph{Computers \& Graphics}, vol.~83, pp. 33--41, 2019.

\bibitem{lin2019site}
Y.~Lin, P.~Peers, and A.~Ghosh, ``On-site example-based material appearance
  acquisition,'' in \emph{Computer Graphics Forum}, vol.~38, no.~4.\hskip 1em
  plus 0.5em minus 0.4em\relax Wiley Online Library, 2019, pp. 15--25.

\bibitem{hertzmann2005example}
A.~Hertzmann and S.~M. Seitz, ``Example-based photometric stereo: Shape
  reconstruction with general, varying brdfs,'' \emph{IEEE Transactions on
  Pattern Analysis and Machine Intelligence}, vol.~27, no.~8, pp. 1254--1264,
  2005.

\bibitem{hertzmann2003shape}
------, ``Shape and materials by example: A photometric stereo approach,'' in
  \emph{2003 IEEE Computer Society Conference on Computer Vision and Pattern
  Recognition, 2003. Proceedings.}, vol.~1.\hskip 1em plus 0.5em minus
  0.4em\relax IEEE, 2003, pp. I--I.

\bibitem{ikeuchi1981determining}
K.~Ikeuchi, ``Determining surface orientations of specular surfaces by using
  the photometric stereo method,'' \emph{IEEE Transactions on Pattern Analysis
  and Machine Intelligence}, no.~6, pp. 661--669, 1981.

\bibitem{dong2010manifold}
Y.~Dong, J.~Wang, X.~Tong, J.~Snyder, Y.~Lan, M.~Ben-Ezra, and B.~Guo,
  ``Manifold bootstrapping for svbrdf capture,'' \emph{ACM Transactions on
  Graphics (TOG)}, vol.~29, no.~4, pp. 1--10, 2010.

\bibitem{guarnera2016brdf}
D.~Guarnera, G.~C. Guarnera, A.~Ghosh, C.~Denk, and M.~Glencross, ``Brdf
  representation and acquisition,'' in \emph{Computer Graphics Forum}, vol.~35,
  no.~2.\hskip 1em plus 0.5em minus 0.4em\relax Wiley Online Library, 2016, pp.
  625--650.

\bibitem{dong2019deep}
Y.~Dong, ``Deep appearance modeling: A survey,'' \emph{Visual Informatics},
  vol.~3, no.~2, pp. 59--68, 2019.

\bibitem{deschaintre2020guided}
V.~Deschaintre, G.~Drettakis, and A.~Bousseau, ``Guided fine-tuning for
  large-scale material transfer,'' in \emph{Computer Graphics Forum}, vol.~39,
  no.~4.\hskip 1em plus 0.5em minus 0.4em\relax Wiley Online Library, 2020, pp.
  91--105.

\bibitem{deschaintre2018single}
V.~Deschaintre, M.~Aittala, F.~Durand, G.~Drettakis, and A.~Bousseau,
  ``Single-image svbrdf capture with a rendering-aware deep network,''
  \emph{ACM Transactions on Graphics (TOG)}, vol.~37, no.~4, pp. 1--15, 2018.

\bibitem{li2017modeling}
X.~Li, Y.~Dong, P.~Peers, and X.~Tong, ``Modeling surface appearance from a
  single photograph using self-augmented convolutional neural networks,''
  \emph{ACM Transactions on Graphics (ToG)}, vol.~36, no.~4, pp. 1--11, 2017.

\bibitem{ye2018single}
W.~Ye, X.~Li, Y.~Dong, P.~Peers, and X.~Tong, ``Single image surface appearance
  modeling with self-augmented cnns and inexact supervision,'' in
  \emph{Computer Graphics Forum}, vol.~37, no.~7.\hskip 1em plus 0.5em minus
  0.4em\relax Wiley Online Library, 2018, pp. 201--211.

\bibitem{nam2016simultaneous}
G.~Nam, J.~H. Lee, H.~Wu, D.~Gutierrez, and M.~H. Kim, ``Simultaneous
  acquisition of microscale reflectance and normals,'' \emph{ACM Transactions
  on Graphics (TOG)}, vol.~35, no.~6, pp. 1--11, 2016.

\bibitem{merzbach2017high}
S.~Merzbach, M.~Weinmann, and R.~Klein, ``High-quality multi-spectral
  reflectance acquisition with x-rite tac7,'' in \emph{Proceedings of the
  Workshop on Material Appearance Modeling}, 2017, pp. 11--16.

\bibitem{photoptics19}
R.~Alcain., C.~Heras., I.~Salinas., J.~López., and C.~Aliaga., ``Microscale
  optical capture system for digital fabric recreation,'' in \emph{Proceedings
  of the 7th International Conference on Photonics, Optics and Laser Technology
  - Volume 1: PHOTOPTICS,}, INSTICC.\hskip 1em plus 0.5em minus 0.4em\relax
  SciTePress, 2019, pp. 114--119.

\bibitem{guo2020materialgan}
Y.~Guo, C.~Smith, M.~Ha{\v{s}}an, K.~Sunkavalli, and S.~Zhao, ``Materialgan:
  reflectance capture using a generative svbrdf model,'' \emph{ACM Transactions
  on Graphics (TOG)}, vol.~39, no.~6, pp. 1--13, 2020.

\bibitem{shorten2019survey}
C.~Shorten and T.~M. Khoshgoftaar, ``A survey on image data augmentation for
  deep learning,'' \emph{Journal of Big Data}, vol.~6, no.~1, p.~60, 2019.

\bibitem{karras2020training}
T.~Karras, M.~Aittala, J.~Hellsten, S.~Laine, J.~Lehtinen, and T.~Aila,
  ``Training generative adversarial networks with limited data,'' \emph{arXiv
  preprint arXiv:2006.06676}, 2020.

\bibitem{chen2020simple}
T.~Chen, S.~Kornblith, M.~Norouzi, and G.~Hinton, ``A simple framework for
  contrastive learning of visual representations,'' in \emph{International
  conference on machine learning}.\hskip 1em plus 0.5em minus 0.4em\relax PMLR,
  2020, pp. 1597--1607.

\bibitem{kauderer2017quantifying}
E.~Kauderer-Abrams, ``Quantifying translation-invariance in convolutional
  neural networks,'' \emph{arXiv preprint arXiv:1801.01450}, 2017.

\bibitem{park2020contrastive}
T.~Park, A.~A. Efros, R.~Zhang, and J.-Y. Zhu, ``Contrastive learning for
  unpaired image-to-image translation,'' \emph{arXiv preprint
  arXiv:2007.15651}, 2020.

\bibitem{zhu2017toward}
J.-Y. Zhu, R.~Zhang, D.~Pathak, T.~Darrell, A.~A. Efros, O.~Wang, and
  E.~Shechtman, ``Toward multimodal image-to-image translation,'' in
  \emph{Advances in Neural Information Processing Systems}, 2017, pp. 465--476.

\bibitem{ronneberger2015u}
O.~Ronneberger, P.~Fischer, and T.~Brox, ``U-net: Convolutional networks for
  biomedical image segmentation,'' in \emph{International Conference on Medical
  image computing and computer-assisted intervention}.\hskip 1em plus 0.5em
  minus 0.4em\relax Springer, 2015, pp. 234--241.

\bibitem{wang2020generalizing}
Y.~Wang, Q.~Yao, J.~T. Kwok, and L.~M. Ni, ``Generalizing from a few examples:
  A survey on few-shot learning,'' \emph{ACM Computing Surveys (CSUR)},
  vol.~53, no.~3, pp. 1--34, 2020.

\bibitem{wang2019few}
T.-C. Wang, M.-Y. Liu, A.~Tao, G.~Liu, B.~Catanzaro, and J.~Kautz, ``Few-shot
  video-to-video synthesis,'' in \emph{Advances in Neural Information
  Processing Systems}, 2019, pp. 5013--5024.

\bibitem{liu2019few}
M.-Y. Liu, X.~Huang, A.~Mallya, T.~Karras, T.~Aila, J.~Lehtinen, and J.~Kautz,
  ``Few-shot unsupervised image-to-image translation,'' in \emph{Proceedings of
  the IEEE International Conference on Computer Vision}, 2019, pp.
  10\,551--10\,560.

\bibitem{isola2017image}
P.~Isola, J.-Y. Zhu, T.~Zhou, and A.~A. Efros, ``Image-to-image translation
  with conditional adversarial networks,'' in \emph{Proceedings of the IEEE
  Conference on Computer Vision and Pattern Recognition}, 2017, pp. 1125--1134.

\bibitem{goodfellow2014generative}
I.~Goodfellow, J.~Pouget-Abadie, M.~Mirza, B.~Xu, D.~Warde-Farley, S.~Ozair,
  A.~Courville, and Y.~Bengio, ``Generative adversarial nets,'' in
  \emph{Advances in Neural Information Processing Systems}, 2014, pp.
  2672--2680.

\bibitem{texler2020arbitrary}
O.~Texler, D.~Futschik, J.~Fi{\v{s}}er, M.~Luk{\'a}{\v{c}}, J.~Lu,
  E.~Shechtman, and D.~S{\`y}kora, ``Arbitrary style transfer using
  neurally-guided patch-based synthesis,'' \emph{Computers \& Graphics},
  vol.~87, pp. 62--71, 2020.

\bibitem{paszke2017automatic}
A.~Paszke, S.~Gross, S.~Chintala, G.~Chanan, E.~Yang, Z.~DeVito, Z.~Lin,
  A.~Desmaison, L.~Antiga, and A.~Lerer, ``Automatic differentiation in
  pytorch,'' 2017.

\bibitem{kingma2014adam}
D.~P. Kingma and J.~Ba, ``Adam: {A} method for stochastic optimization,'' in
  \emph{3rd International Conference on Learning Representations, {ICLR}},
  2015.

\bibitem{castillo2019recent}
C.~Castillo, J.~L{\'o}pez-Moreno, and C.~Aliaga, ``Recent advances in fabric
  appearance reproduction,'' \emph{Computers \& Graphics}, vol.~84, pp.
  103--121, 2019.

\bibitem{kampouris2016fine}
C.~Kampouris, S.~Zafeiriou, A.~Ghosh, and S.~Malassiotis, ``Fine-grained
  material classification using micro-geometry and reflectance,'' in
  \emph{Proceedings of the European Conference on Computer vision
  (ECCV)}.\hskip 1em plus 0.5em minus 0.4em\relax Springer, 2016, pp. 778--792.

\bibitem{theis2015note}
L.~Theis, A.~v.~d. Oord, and M.~Bethge, ``A note on the evaluation of
  generative models,'' \emph{arXiv preprint arXiv:1511.01844}, 2015.

\bibitem{zhou2019context}
Y.~Zhou, X.~Sun, Z.-J. Zha, and W.~Zeng, ``Context-reinforced semantic
  segmentation,'' in \emph{Proceedings of the IEEE Conference on Computer
  Vision and Pattern Recognition}, 2019, pp. 4046--4055.

\bibitem{nielsen2015optimal}
J.~B. Nielsen, H.~W. Jensen, and R.~Ramamoorthi, ``On optimal, minimal brdf
  sampling for reflectance acquisition,'' \emph{ACM Transactions on Graphics
  (TOG)}, vol.~34, no.~6, pp. 1--11, 2015.

\bibitem{deschaintre2019flexible}
V.~Deschaintre, M.~Aittala, F.~Durand, G.~Drettakis, and A.~Bousseau,
  ``Flexible svbrdf capture with a multi-image deep network,'' in
  \emph{Computer Graphics Forum}, vol.~38, no.~4.\hskip 1em plus 0.5em minus
  0.4em\relax Wiley Online Library, 2019, pp. 1--13.

\bibitem{benton2020learning}
G.~Benton, M.~Finzi, P.~Izmailov, and A.~G. Wilson, ``Learning invariances in
  neural networks,'' \emph{arXiv preprint arXiv:2010.11882}, 2020.

\bibitem{antoniou2017data}
A.~Antoniou, A.~Storkey, and H.~Edwards, ``Data augmentation generative
  adversarial networks,'' \emph{arXiv preprint arXiv:1711.04340}, 2017.

\bibitem{sandfort2019data}
V.~Sandfort, K.~Yan, P.~J. Pickhardt, and R.~M. Summers, ``Data augmentation
  using generative adversarial networks (cyclegan) to improve generalizability
  in ct segmentation tasks,'' \emph{Scientific reports}, vol.~9, no.~1, pp.
  1--9, 2019.

\bibitem{georgiev2018arnold}
I.~Georgiev, T.~Ize, M.~Farnsworth, R.~Montoya-Vozmediano, A.~King, B.~V.
  Lommel, A.~Jimenez, O.~Anson, S.~Ogaki, E.~Johnston \emph{et~al.}, ``Arnold:
  A brute-force production path tracer,'' \emph{ACM Transactions on Graphics
  (TOG)}, vol.~37, no.~3, pp. 1--12, 2018.

\bibitem{cook1982reflectance}
R.~L. Cook and K.~E. Torrance, ``A reflectance model for computer graphics,''
  \emph{ACM Transactions on Graphics (ToG)}, vol.~1, no.~1, pp. 7--24, 1982.

\bibitem{gao2019deep}
D.~Gao, X.~Li, Y.~Dong, P.~Peers, K.~Xu, and X.~Tong, ``Deep inverse rendering
  for high-resolution svbrdf estimation from an arbitrary number of images,''
  \emph{ACM Transactions on Graphics (TOG)}, vol.~38, no.~4, pp. 1--15, 2019.

\bibitem{weinmann-2014-materialclassification}
M.~Weinmann, J.~Gall, and R.~Klein, ``Material classification based on training
  data synthesized using a btf database,'' in \emph{Computer Vision - ECCV 2014
  - 13th European Conference, Zurich, Switzerland, September 6-12, 2014,
  Proceedings, Part III}.\hskip 1em plus 0.5em minus 0.4em\relax Springer
  International Publishing, 2014, pp. 156--171.

\end{thebibliography}

\begin{IEEEbiography}[{\includegraphics[width=1in,height=1.25in,clip,keepaspectratio]{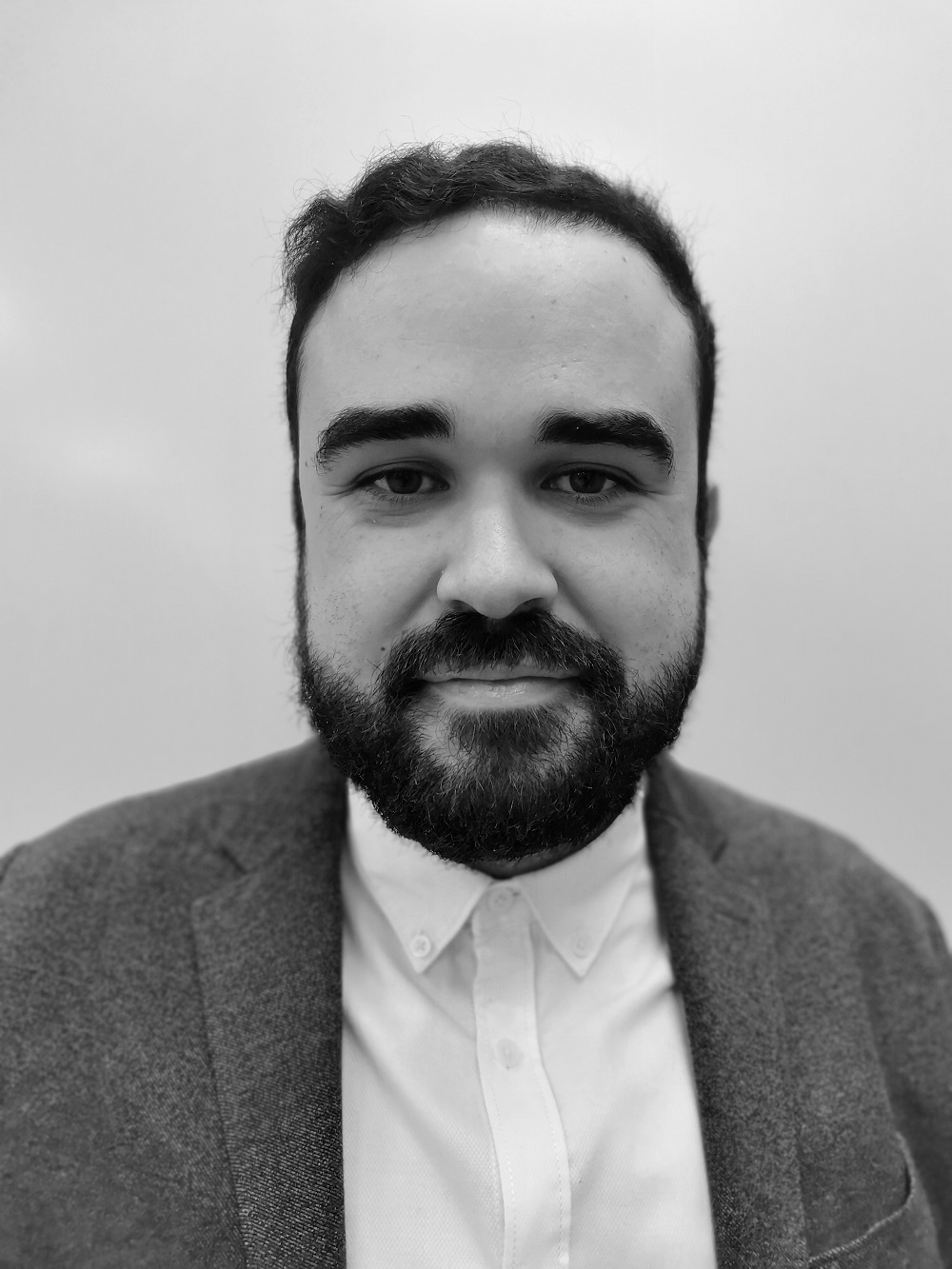}}]{Carlos Rodriguez - Pardo}
 is a research engineer at SEDDI and a PhD student at the Universidad Carlos III de Madrid, Spain (UC3M). His research interests include computer vision and artificial intelligence. In 2018, he was awarded a distinction at the MSc in Artificial Intelligence at the University of Edinburgh. He completed a double BSc degree in Computer Science and Business Administration (UC3M) in 2017. He was a researcher at the Applied Artificial Intelligence Group (UC3M), working in AR applications (2013) and in data science problems (2016-2017). Carlos has served as a reviewer to conferences and journals, such as CVPR, ICCV, BMVC, ICLR, or TVCJ.
\end{IEEEbiography}

\begin{IEEEbiography}[{\includegraphics[width=1in,height=1.25in,clip,keepaspectratio]{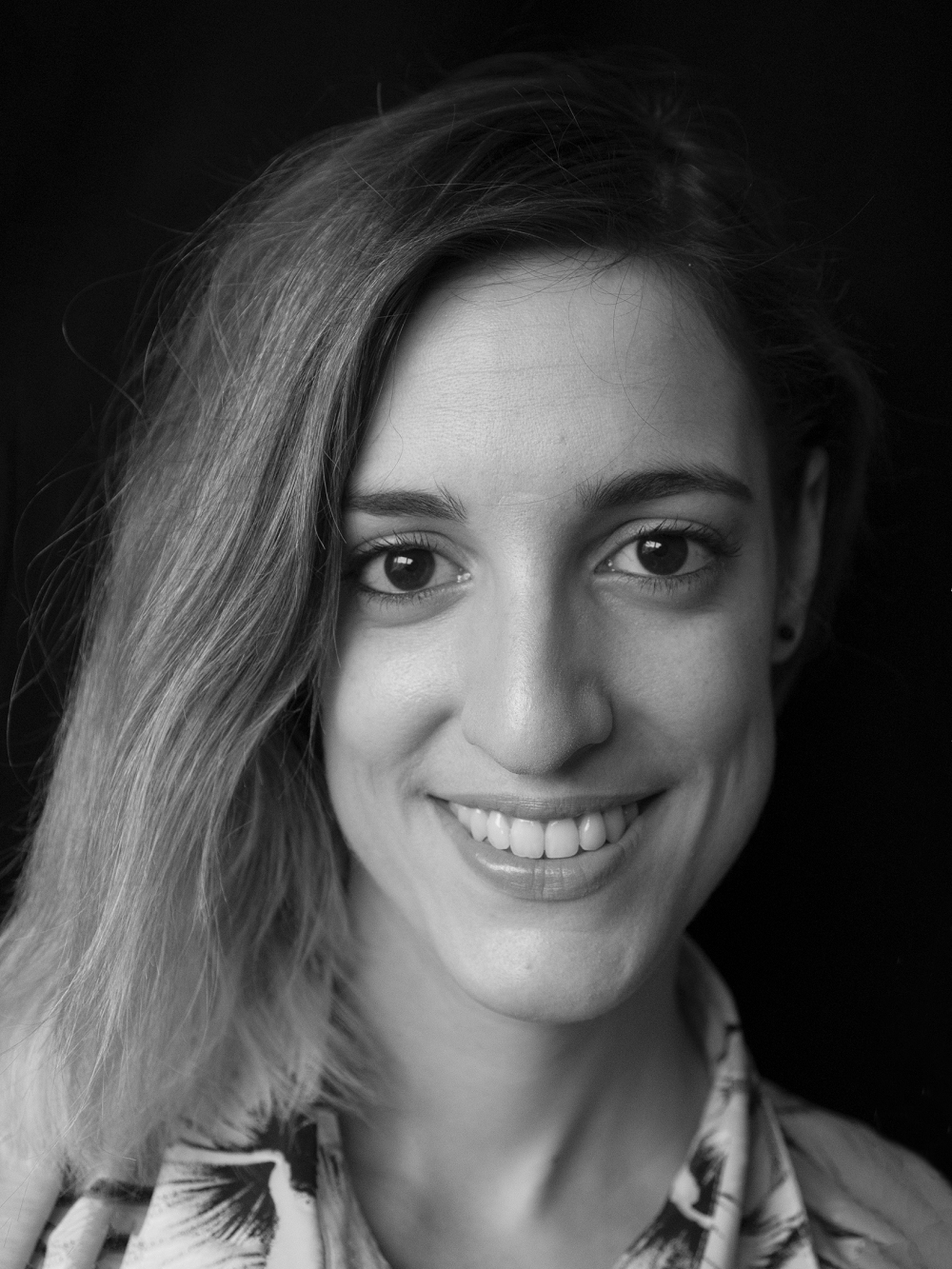}}]{Elena Garces}
received her PhD degree in Computer Science from the University of Zaragoza in 2016. During her PhD studies, she interned twice at the Adobe (San Jose, and Seattle, USA). Her thesis dissertation focused on inverse problems of appearance capture, intrinsic decomposition from single images, video, and lightfields. She was post-doctoral researcher (2016-2018) at Technicolor R\&D (Rennes, France) working on lightfields processing, and post-doctoral Juan de la Cierva Fellow (2018-2019) at the Multimodal Simulation Lab (URJC). Since 2019 she is Senior Research Scientist at SEDDI, leading the optical capture and rendering teams. She has published over 15 papers in top-tier conferences in the areas of computer graphics, vision, and machine learning, as well as authored six patents. Elena serves regularly as reviewer or PC-Member in top-tier computer vision and graphics conferences and journals such as SIGGRAPH, CVPR, ICCV, IJCV, TVCG, or EGSR. 
\end{IEEEbiography}

\end{document}